\documentclass{article}
\usepackage{jfrExamplee}
\usepackage{graphicx}
\usepackage{apalike}
\usepackage{setspace}

\usepackage{comment}
\usepackage{adjustbox}
\usepackage{algorithm}
\usepackage{algpseudocode}
\usepackage{booktabs}
\usepackage{graphicx}
\usepackage{tabularx}
\usepackage{amsmath}
\usepackage{amssymb}
\usepackage{url}
\usepackage{subcaption}
\usepackage{color}
\usepackage{multirow}
\usepackage{multicol}
\usepackage{todonotes}

\newcommand{\R}[1]{\mathbb{R}^{#1}}
\newcommand{\RR}[2]{\mathbb{R}^{#1 \times #2}}

\newcommand{\refEq}[1]{(\ref{#1})}

\def\bfc{{\boldsymbol{c}}}

\def\bfB{{\boldsymbol{B}}}

\def\bfD{{\boldsymbol{D}}}

\def\bfR{{\boldsymbol{R}}}
\def\bfS{{\boldsymbol{S}}}
\def\bfT{{\boldsymbol{T}}}

\def\bfW{{\boldsymbol{W}}}

\def\bfZ{{\boldsymbol{Z}}}

\def\bfzero{{\boldsymbol{0}}}
\def\bfone{{\boldsymbol{1}}}
\def\half{\frac{1}{2}~}

\newcolumntype{P}[1]{>{\centering\arraybackslash}p{#1}}

\newcommand{\changes}[1]{\textcolor{black}{#1}}
\newcommand{\secondchanges}[1]{\textcolor{black}{#1}}
\newcommand{\thirdchanges}[1]{\textcolor{black}{#1}}

\title{Semantic keypoint-based pose estimation from single RGB frames}

\author{

Karl Schmeckpeper\thanks{karls@seas.upenn.edu, University of Pennsylvania, 3330 Walnut Street, Philadelphia, PA, USA 19104}, 
Philip R. Osteen\thanks{philip.r.osteen.civ@mail.mil, DEVCOM Army Research Laboratory, 2800 Powder Mill Road, Adelphi, MD, USA 20783},
Yufu Wang\footnotemark[1], Georgios Pavlakos\thanks{University of California, Berkeley, 253 Cory Hall, Berkeley, CA 94720. Work done while at the University of Pennsylvania}, Kenneth Chaney\footnotemark[1], \\

\textbf{Wyatt Jordan}\footnotemark[2],
\textbf{Xiaowei Zhou}\thanks{Zhejiang University, 866 Yuhangtang Rd, Hangzhou, China. Work done while at the University of Pennsylvania},
\textbf{Konstantinos G. Derpanis}\thanks{Ryerson University, 350 Victoria Street, Toronto, ON, Canada},
\textbf{Kostas Daniilidis}\footnotemark[1]
}

\begin{document}
\maketitle

\begin{abstract}
This paper presents an approach to estimating the continuous
\thirdchanges{6-DoF pose} of an object from a single RGB image. The  approach combines semantic keypoints predicted by a convolutional network (convnet) with a deformable shape model.
Unlike prior 
\thirdchanges{investigators}, we are agnostic to whether the object is textured or textureless, as the convnet learns the optimal representation from the available \thirdchanges{training-image} data. Furthermore, the approach can be applied to instance- and class-based pose recovery. 
Additionally, we accompany our main pipeline with a technique for \changes{semi-}automatic data generation from unlabeled videos.
This \thirdchanges{procedure} allows us to train the learnable components of our method with minimal manual intervention \thirdchanges{in the labeling process}.
%
Empirically, we show that
\thirdchanges{our} approach can accurately recover the 6-DoF object pose for both instance- and class-based scenarios
\thirdchanges{even against} a cluttered background. 
We apply our approach \secondchanges{both to several}, existing, large-scale datasets \thirdchanges{-} \changes{including} PASCAL3D+, \changes{LineMOD-Occluded, YCB-Video, and TUD-Light} \thirdchanges{-} and, using our labeling pipeline, to \changes{a} new dataset with novel object classes \changes{\thirdchanges{that} we introduce here}.
\thirdchanges{Extensive empirical evaluations show that}
\changes{our approach is able to provide pose estimation results comparable to the state of the art}.
\end{abstract}

\section{Introduction}
\begin{figure}
  \centering
  \begin{subfigure}[b]{0.2\textwidth}
    \includegraphics[width=\linewidth]{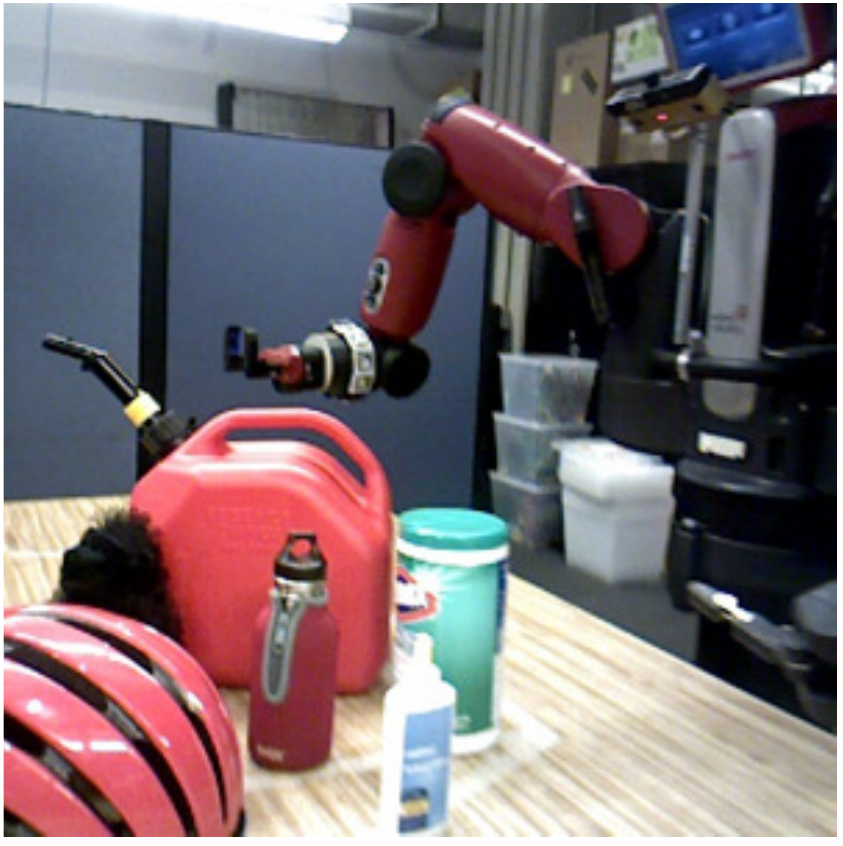}
  \end{subfigure}
  \begin{subfigure}[b]{0.2\textwidth}
  \includegraphics[width=\linewidth]{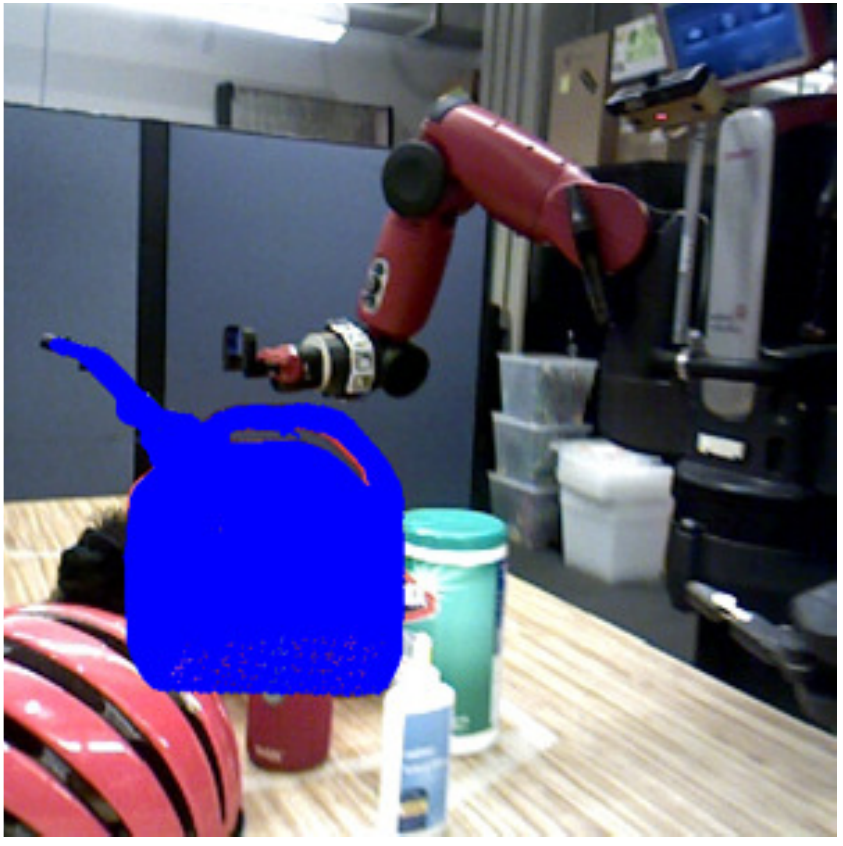}
  \end{subfigure}  
  \begin{subfigure}[b]{0.2\textwidth}
  \includegraphics[width=\linewidth]{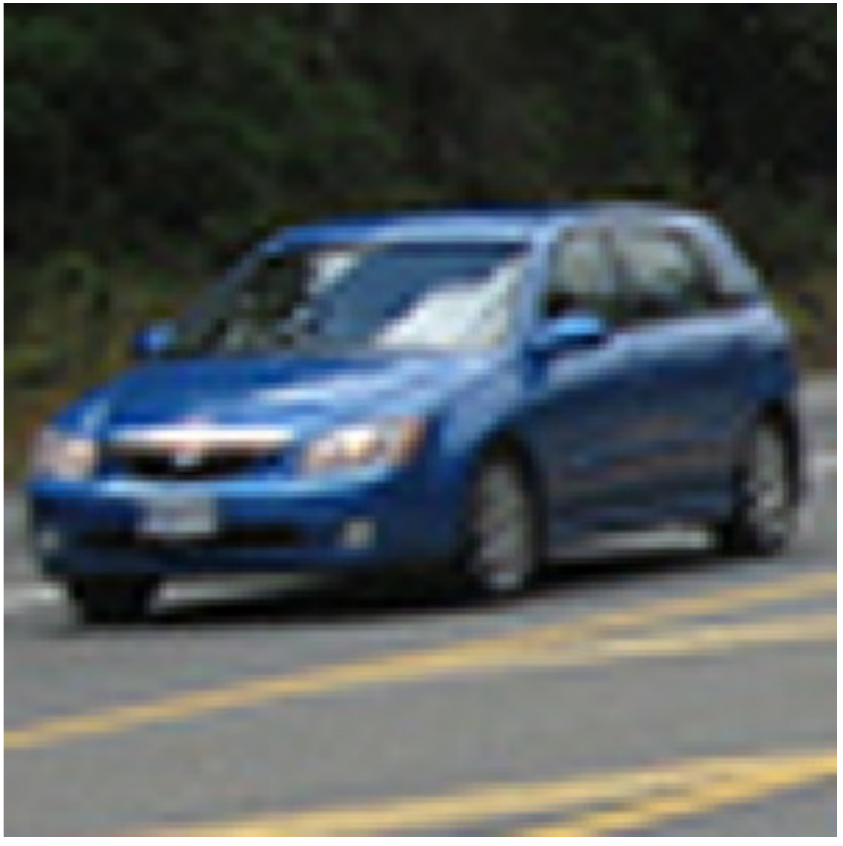}
  \end{subfigure}
  \begin{subfigure}[b]{0.2\textwidth}
  \includegraphics[width=\linewidth]{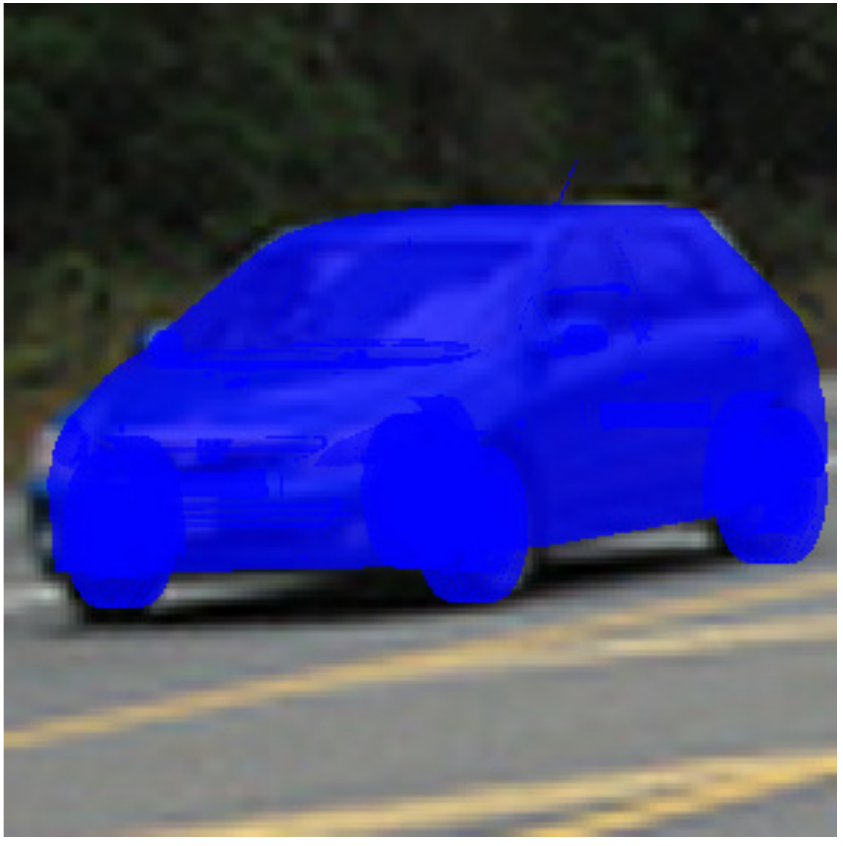}
 \end{subfigure}  
  \caption{Overview of approach. Given a single RGB image of an object (\thirdchanges{first and third columns}), we estimate its 6-DoF pose. 
The corresponding CAD model is overlaid on the image (second and fourth columns) using the estimated pose. 
Our method
deals with both instance-based (left \thirdchanges{pair}) and class-based scenarios (right \thirdchanges{pair}).
}\label{fig:intro}
\end{figure}

This paper addresses the task of estimating 
\thirdchanges{an object's} continuous, six degrees-of-freedom (6-DoF) pose (3D translation and rotation) of an object from a single image.  Despite its importance in a variety of applications, e.g., robotic manipulation, navigation, etc., and its intense study, most solutions tend to treat objects on a case-by-case basis.  
\thirdchanges{For instance, approaches can be distinguished by whether they apply to ``sufficiently'' textured objects or apply to textureless objects.}
\thirdchanges{In addition, some} approaches focus on instance-based object detection, while others address object classes. In this work we strive for an approach where the admissibility of objects considered is as wide as possible (examples in Figure~\ref{fig:intro}).

\begin{figure*}
  \centering
  \begin{subfigure}[b]{0.2\textwidth}
    \includegraphics[width=\linewidth]{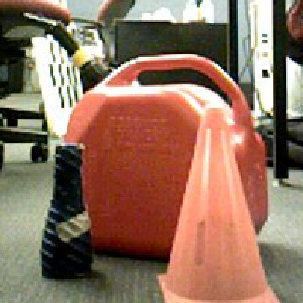}
    \caption{}
  \end{subfigure}
  \begin{subfigure}[b]{0.2\textwidth}
  \includegraphics[width=\linewidth]{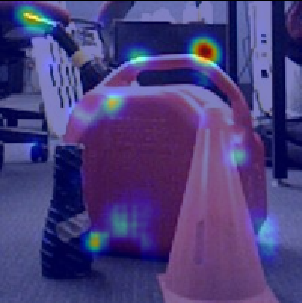}
  \caption{}
  \end{subfigure}  
  \begin{subfigure}[b]{0.2\textwidth}
  \includegraphics[width=\linewidth]{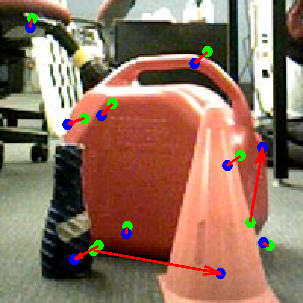}
  \caption{}
  \end{subfigure}
  \begin{subfigure}[b]{0.2\textwidth}
  \includegraphics[width=\linewidth]{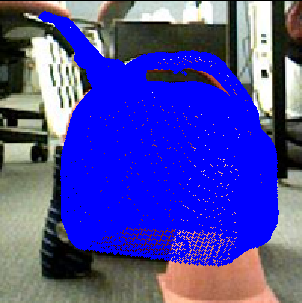}
  \caption{}
 \end{subfigure}  
  \caption{Processing pipeline of our approach. Given a single RGB image of an object (a), we localize a set of class-specific keypoints using a convnet. The output of this step is a set of heatmaps\secondchanges{, one} for each keypoint, which are combined for visualization in (b), sometimes leading to false detections. In (c), green dots represent the detected keypoints and the corresponding blue dots (connected with an arrow) the groundtruth locations. For robustness against such localization errors, we solve a fitting problem to enforce global consistency of the keypoints, where the response of the heatmaps is used as a measure of certainty for each keypoint. The optimization recovers the full 6-DoF pose of the object (d).}
  \label{fig:pipeline}
\end{figure*}

Our approach combines statistical models of appearance and the 3D shape layout of objects for pose estimation. It consists of two stages that first reason about the 2D projected shape of an object, as captured by a set of 2D semantic keypoints, then estimates the 3D pose consistent with those keypoints \thirdchanges{(Figure~\ref{fig:pipeline})}.
In the first stage, we use a high capacity, convolutional network (convnet) to predict a set of semantic keypoints. Here, the network takes advantage of its ability to aggregate appearance information over a wide \thirdchanges{receptive field}, as compared to localized part models, e.g., \cite{gu2010}, to make reliable predictions of the semantic keypoints.  In the second stage, the semantic keypoint predictions are used to reason explicitly about the intra-class shape variability and the camera pose modeled by a weak- or full-perspective camera model. Pose estimates are realized by maximizing the geometric consistency between the parametrized, deformable model and the 2D semantic keypoints. While this work focuses on RGB-based pose estimation, our method can provide a robust way to initialize the iterative closest point (ICP) algorithm \cite{ICP} to further refine the pose in the case where a corresponding point cloud is provided with the image.

In this paper, we \thirdchanges{additionally focus on} \changes{efficiently} collecting the data to train the learnable components of our approach.
More specifically, \thirdchanges{although} the convnet used for the detection of the semantic keypoints requires a relatively small number of images for training, 
\thirdchanges{annotation time can become excessive as all relevant keypoints must be identified.}
To bypass the manual labor, we propose a \changes{semi-}automatic procedure that can greatly accelerate deployment of our algorithm, particularly in the instance-based scenario.
Our method relies on 3D reconstruction of the object from an unlabeled input video, and selection of a small number of keypoints on the final 3D surface. Projecting the 3D keypoints on all the images of the video allows us to generate a large number of training data for the keypoint localization network. This procedure can be repeated for multiple videos 
\thirdchanges{to significantly increase the quantity of data available}
for training, capturing multiple backgrounds, poses, and lighting conditions.

In summary, this paper extends our previous work published in ICRA 2017~\cite{Pavlakos2017}.
Our main contribution lies in presenting an accurate and robust approach for 6-DoF object pose estimation, both for the class-based and the instance-based cases.
Here we extend our previous work
in the following ways.
\begin{itemize}
    \item We present an approach for semi-automatic data generation which is appropriate to train the learnable components of the original method
    \item We release a dataset of outdoor pose estimation challenges labeled with this pipeline
    \item We carefully evaluate the accuracy of the data labeling and the effect it has on the rest of the pipeline
    \item We provide additional comparisons \thirdchanges{on new datasets} between our pose estimation algorithm and those developed since the release of the original paper, demonstrating that our approach has withstood the test of time.
\end{itemize}

\section{Related Works}

\label{sec:related_works}
Estimating the 6-DoF pose of an object from a single image has attracted significant study. Given a rigid 3D object model and a set of 2D-to-3D point correspondences, various solutions have been explored, e.g., \cite{fischler1981,lepetit2009}. This is commonly referred to as the Perspective-n-Point problem (PnP). To relax the assumption of known 2D landmarks, a number of approaches have considered the detection of discriminative image keypoints \cite{collet2009,collet2011,xie2013}, such as SIFT \cite{lowe2004}, with highly textured objects.  A drawback with these approaches is that they are inadequate for addressing textureless objects and  their performance is susceptible to scene clutter. An alternative to sparse discriminative keypoints is offered by dense methods~\cite{brachmann2014learning,brachmann2016uncertainty,doumanoglou2016recovering,michel2016global}, where every pixel or patch is voting for the object pose. These approaches are also applicable for textureless objects, however, the assumption that a corresponding instance-specific 3D model is available for each object limits their general applicability.

Holistic template-based approaches are one of the earliest approaches considered in the object detection literature. To accommodate appearance variation due to camera capture viewpoint, a set of template images of the object instance are captured about the view sphere and are compared to the input image at runtime. In recent years, template-based methods have received renewed interest due to the advent of accelerated matching schemes and their ability to detect textureless objects by way of focusing their model description on the object shape \cite{muja2011,Hinterstoisser2012,rios2013,xie2013,cao2016}. While impressive results in terms of accuracy and speed have been demonstrated, holistic template-based approaches are limited to instance-based object detection and are not robust to occlusions. To address class variability and viewpoint, various approaches have used a collection of 2D appearance-based part templates trained separately on discretized views \cite{gu2010,fidler2012,pepik2012,xiang2014,zhu2014single}.  

Convolutional networks \secondchanges{or} convnets \cite{lecun1989,krizhevsky2012} have emerged as the method of choice for a variety of vision problems.  Closest to the current work is their application in camera viewpoint and keypoint prediction. Convnets have been used to predict the camera's viewpoint with respect to the object by way of direct regression or casting the problem as classification into a set discrete views \cite{massa2014,tulsiani2015vk,su2015}.  While these approaches allow for object category pose estimation they do not provide fine-grained information about the 3D layout of the object. Convnet-based keypoint prediction for human pose estimation (e.g., \cite{toshev2014,zhou2015sparseness,newell2016stacked,wei2016cpm}) has attracted considerable study, while limited attention has been given to their application with generic object categories \cite{long2014,tulsiani2015vk}. Their success is due in part to the high discriminative capacity of the network.  Furthermore, their ability to aggregate information over a wide field of view allows for the resolution of ambiguities (e.g., symmetry) and for localizing occluding joints.

Statistical shape-based models tackle recognition by aligning a shape subspace model to image features.  While originally proposed in the context of 2D shape  \cite{cootes1995} they have proven useful for modelling the 3D shape of a host of object classes, e.g., faces \cite{cao20133d}, cars \cite{zia2013detailed,murthy2016reconstructing}, and human pose \cite{ramakrishna2012}.  In \cite{zhu_popup}, data-driven discriminative landmark hypotheses were combined with a 3D deformable shape model and a weak perspective camera model in a convex optimization framework to globally recover the shape and pose of an object in a single image.  Here, we adapt this approach and extend it with a perspective camera model, in cases where the camera intrinsics are known.

Since our initial work on pose estimation from semantic keypoints \cite{Pavlakos2017}, several other works have been inspired by or have used our method.
Wang et al. demonstrate the effectiveness of using learned keypoints for object tracking \cite{wang20196}.
Qin et al. show a pipeline that learns task specific keypoints for manipulation \cite{qin2019keto}.
Manuelli et al. use semantic keypoints as their object representation for manipulation \cite{manuelli2019kpam}.
Peng et al. propose a refinement of the keypoint detection for pose estimation using pixel-wise voting \cite{peng2019pvnet}.
Hu et al. use object segmentations rather than object bounding boxes to input detected objects into their semantic keypoint based pose estimation pipeline \cite{Hu_2019_CVPR}.
Zhou et al. learn a set of category agnostic keypoints for object pose estimation \cite{zhou2018starmap}.

Other works directly use our work or similar approaches.
Zuo et al. show that a keypoint based pose estimation pipeline can be used to track the position of a low cost robotic arm that does not have its own sensors, enabling cheap manipulation \cite{zuo2019craves}.
The keypoint-based pose estimation pipeline was used by \cite{bowman2017probabilistic} as the first step in their semantic \secondchanges{SLAM} pipeline.
Our pipeline has been used by \cite{vasilopoulos2018reactive,vasilopoulos2020reactive,vasilopoulos_ral_2020} to detect and localize objects for reactive planning for navigation, and was used by the Robotics Collaborative Techonology Alliance (RCTA) program to rapidly collect and annotate data for pose estimation for mobile manipulation~\cite{spie_perception,spie_t3}.

\changes{RGB-based instance-level pose estimation has also made improvements since our method was initially proposed, largely due to deep network based pipelines~\cite{kehl2017ssd,xiang2017posecnn,mousavian20173d,kehl2017ssd,tekin2018real,tremblay2018deep,do2018deep,li2018deepim,montserrat2019multi,labbe2020cosypose,ke2020gsnet,hou2020mobilepose} and higher quality training data~\cite{zeng2017multi,hodan2018bop,hodan2020bop,sundermeyer2020augmented,wang2020self6d}. Interestingly, while several single-shot methods have focused on speed~\cite{kehl2017ssd,tekin2018real,hou2020mobilepose}, many state of the art approaches still follow the two-stage approach, where 2D-3D correspondences are detected with a deep network in the first stage, and 6D pose is solved in the second stage with PnP methods. Different intermediate representations have been explored to establish correspondences, such as 3D bounding box corners~\cite{rad2017bb8}, dense object coordinates~\cite{park2019pix2pose}, dense fragments~\cite{hodan2020epos}, pixel-wise voting~\cite{peng2019pvnet} and their hybrids~\cite{song2020hybridpose}. We will show through new experiments in Section 6.2 that our simple semantic keypoint based representation is still effective and competitive, even in challenging occluded cases.}






\section{Pose estimation from Semantic Keypoints}
\thirdchanges{Our} pipeline includes object detection, keypoint localization, and pose optimization. As object detection has been a well studied problem, we assume that a bounding box around the object has been provided by an off-the-shelf object detector, e.g., Faster R-CNN \cite{ren2015faster}, and focus on 
\thirdchanges{the later two processes.}

\subsection{Pose optimization}\label{sec:optimization}
Given the keypoint locations on the 3D model as well as their correspondences in the 2D image, one naive approach is to simply apply an existing PnP algorithm to solve for the 6-DoF pose. This approach is problematic because the keypoint predictions can be rendered imprecise due to occlusions and false detections in the background. Moreover, the exact 3D model of the object instance in the testing image is often unavailable. To address these difficulties, we 
fit a deformable shape model to the 2D detections 
while considering the uncertainty in keypoint predictions.   \secondchanges{This approach additionally allows the optimzation to include the uncertainty of the detections, making it better able to handle false or imprecise detections.}

We build a deformable shape model for each object category using 3D CAD models with annotated keypoints. More specifically, the $p$ keypoint locations on a 3D object model are denoted by $\bfS\in\RR{3}{p}$ and
\begin{align}\label{eq:shape-model}
    \bfS = \bfB_0 + \sum_{i=1}^{k} c_i\bfB_i,
\end{align}
where $\bfB_0$ is the mean shape of the given 3D model and $\bfB_1,\dotsc,\bfB_k$ are several modes of possible shape variability computed by Principal Component Analysis (PCA).

Given detected keypoints in an image, which are denoted by $\bfW\in\RR{2}{p}$, the goal is to estimate the rotation $\bfR\in\RR{3}{3}$ and translation $\bfT\in\RR{3}{1}$ between the object  and camera frames as well as the coefficients of the shape deformation $\bfc=[c_1,\cdots,c_k]^\top$. 

The inference is formulated as the following optimization problem:
\begin{align}\label{eq:cost}
    \min_{\theta} ~~ & \half \left\| \xi(\theta)\bfD^{\frac{1}{2}} \right\|_F^2 + \frac{\lambda}{2} \|\bfc\|_2^2,
\end{align}
where $\theta$ is the set of unknowns, $\xi(\theta)$ denotes the fitting residuals dependent on $\theta$, 
and the Tikhonov regularizer $\|\bfc\|_2^2$ is introduced to penalize large deviations from the mean shape. 

To incorporate the uncertainty in 2D keypoint predictions, a diagonal weighting matrix $\bfD\in\RR{p}{p}$ is introduced:
\begin{align}
\bfD &= \begin{bmatrix}
            d_1 & 0 & \cdots & 0\\
            0 & d_2 & \cdots & 0\\
            \vdots & \vdots & \ddots & \vdots\\
            0 & 0 & \cdots & d_p\\
            \end{bmatrix},
\end{align}
where $d_{i}$ indicates the localization confidence of the $i$th keypoint in the image. In our implementation, $d_{i}$ is assigned the peak value in the heatmap corresponding to the $i$th keypoint. As shown previously \cite{newell2016stacked}, the peak intensity of the heatmap provides a good indicator for the visibility of a keypoint in the image.

The fitting residuals, $\xi(\theta)$, measure the differences between the given 2D keypoints, provided by the previous processing stage, and the projections of 3D keypoints. Two camera models are next considered.

\subsubsection{Weak perspective model}

If the camera intrinsic parameters are unknown, the weak perspective camera model is adopted, which is usually a good approximation to the full perspective case when the camera is relatively far away from the object. In this case, the reprojection error is written as 
\begin{align}
    \xi(\theta) = \bfW - s \bar{\bfR}\left(\bfB_0 + \sum_{i=1}^{k} c_i\bfB_i\right) - \bar{\bfT}\bfone^\top,
\end{align}
where $s$ is a scalar, $\bar{\bfR}\in\RR{2}{3}$ and $\bar{\bfT}\in\R{2}$ denote the first two rows of $\bfR$ and $\bfT$, respectively, and $\theta=\{s,\bfc,\bar{\bfR},\bar{\bfT}\}$.

The problem in \refEq{eq:cost} is continuous and in principal can be locally solved by any gradient-based method. We solve it with a block coordinate descent scheme because of its fast convergence and the simplicity in implementation. We alternately update each of the variables while fixing the others. The updates of $s$, $\bfc$ and $\bar{\bfT}$ are simply solved using closed-form least squares solutions. The update of $\bar{\bfR}$ should consider the $SO(3)$ constraint. Here, the Manopt toolbox \cite{boumal2014manopt} is used to optimize $\bar{\bfR}$ over the Stiefel manifold. As the problem in \refEq{eq:cost} is non-convex, we further adopt a convex relaxation approach \cite{zhou20153d} to initialize the optimization. More specifically, we only estimate the pose parameters while fixing the 3D model as the mean shape in the initialization stage. By setting $\bfc=\bfzero$ and replacing the orthogonality constraint on $\bar{\bfR}$ by the spectral norm regularizer, the problem in \refEq{eq:cost} can be converted to a convex program and solved with global optimality \cite{zhou20153d}.

\subsubsection{Full perspective model}

If the camera intrinsic parameters are known, the full perspective camera model is used, and the residuals are defined as
\begin{align}
\xi(\theta) = \tilde{\bfW}\bfZ- {\bfR}\left(\bfB_0 + \sum_{i=1}^{k} c_i\bfB_i\right) - {\bfT}\bfone^\top,
\end{align}
where $\tilde{\bfW}\in\RR{3}{p}$ represents the normalized homogeneous coordinates of the 2D keypoints and $\bfZ$ is a diagonal matrix: 
\begin{align}
\bfZ &= \begin{bmatrix}
            z_1 & 0 & \cdots & 0\\
            0 & z_2 & \cdots & 0\\
            \vdots & \vdots & \ddots & \vdots\\
            0 & 0 & \cdots & z_p\\
            \end{bmatrix},
\end{align}
where $z_i$ is the depth for the $i$th keypoint in 3D. Intuitively, the distances from the 3D points to the rays crossing the corresponding 2D points are minimized. In this case, the unknown parameter set $\theta$ is given by $\{\bfZ,\bfc,\bfR,\bfT\}$.

The optimization here is similar to the alternating scheme in the weak perspective case. The update of $\bfZ$ also admits a closed-form solution and the update of $\bfR$ can be analytically solved by the orthogonal Procrustes analysis. To avoid local minima, the optimization is initialized by the weak perspective solution.  

\subsection{Keypoint localization}
The keypoint localization step employs the ``stacked hourglass'' network architecture~\cite{newell2016stacked} that has been shown to be particularly effective for 2D human pose estimation. Motivated by this success, we use the same network design and train the network for object keypoint localization. 

\begin{figure*}
  \centering
  \includegraphics[width=0.8\linewidth,trim={5cm 5cm 5cm 5cm},clip]{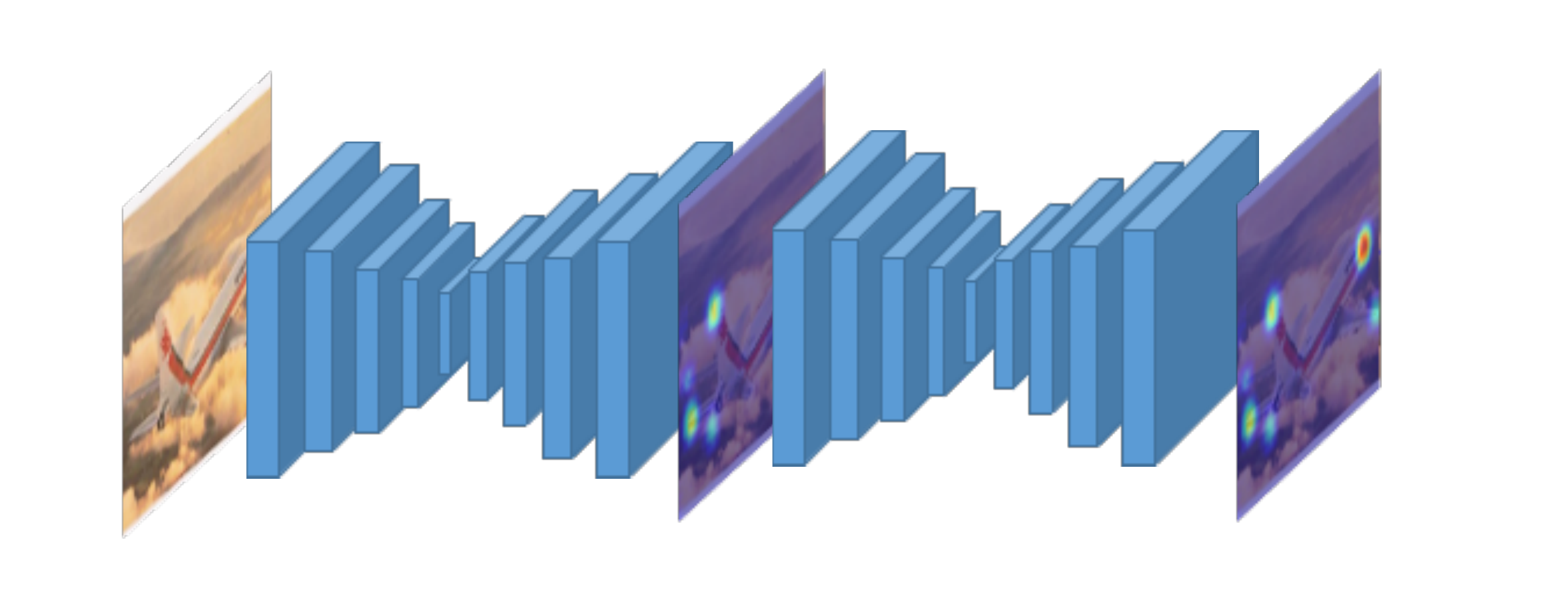}\\
  \vspace{-2em}
	  \begin{tabular}{P{4.5cm} P{4.5cm} P{5cm}} 
	  \scriptsize{Image} & \scriptsize{Intermediate heatmaps} & \scriptsize{Output heatmaps} \\ 
	  \end{tabular}
  \caption{Overview of the stacked hourglass architecture~\cite{newell2016stacked}. Here, two hourglass modules are stacked together. The symmetric nature of the design allows for bottom-up processing (from high to low resolution) in the first half of the module, and top-down processing (from low to high resolution) in the second half. Intermediate supervision is applied after the first module. The heatmap responses of the second module represent the final output of the network that is used for keypoint localization.}\label{fig:hourglass}
\vspace{-1em}
\end{figure*}

\vspace{5pt}\noindent{\bf Network architecture }
A high level overview of the main network components is presented in Figure~\ref{fig:hourglass}. The network takes as input an RGB image, and outputs a set of heatmaps, one per keypoint, with the intensity of the heatmap indicating the confidence of the respective keypoint to be located at this position. The network consists of two hourglass components, where each component can be further subdivided into two main processing stages. In the first stage, a series of convolutional and max-pooling layers are applied to the input. After each max-pooling layer, the resolution of the feature maps decreases by a factor of two, allowing the next convolutional layer to process the features at a coarser scale. This sequence of processing continues until reaching the lowest resolution ($4\times4$ feature maps), which is illustrated by the smallest layer in the middle of each module in Figure~\ref{fig:hourglass}. Following these downsampling layers, the processing continues with a series of convolutional and \secondchanges{nearest-neighbor} upsampling layers. Each upsampling layer increases the resolution by a factor of two. This process culminates with a set of heatmaps at the same resolution as the input of the hourglass module. A second hourglass component is stacked at the end of the first one to refine the output heatmaps. The groundtruth labels used to supervise the training are synthesized heatmaps based on a 2D Gaussian centered at each keypoint with a standard deviation set to one. The $\ell_2$ loss is minimized during training. \secondchanges{Additionally, intermediate supervision is applied} at the end of the first module, which provides a richer gradient signal to the network and guides the learning procedure towards a better optimum \cite{lee2015deeply}.  \secondchanges{We evenly weight the the loss from the intermediate heatmaps with the heatmaps from the last module.  We train the model using the RMSProp optimizer \cite{tieleman2012lecture}.} The heatmap responses of the last module are considered as the final output of the network and the peak in each heatmap indicates the most likely location for the corresponding keypoint. \secondchanges{For more details of our architecture, please see our official implementation.\footnote{Our implementation of the keypoint detector network is available at \url{https://github.com/kschmeckpeper/keypoint-detection}}}

\vspace{5pt}\noindent{\bf Design benefits }
The most critical design element of the hourglass network is the symmetric combination of bottom-up and top-down processing that each hourglass module performs. Given the large appearance changes of objects due to in-class and viewpoint variation, both local and global cues are needed to effectively decide the locations of the keypoints in the image. The consolidation of features across different scales in the hourglass architecture allows the network to successfully integrate both local and global appearance information, and commit to a keypoint location only after this information has been made available to the network. Moreover, the stacking of hourglass modules provides a form of iterative processing that has been shown to be effective with several other recent network designs~\cite{carreira2015iterative,wei2016cpm} and offers additional refinement of the network estimates. Additionally, the application of intermediate supervision at the end of each module has been validated as an effective training strategy, particularly ameliorating the practical issue of vanishing gradients when training a deep neural network \cite{lee2015deeply}. Finally, residual layers are introduced~\cite{he2015residual}, which have achieved state-of-the-art results for many visual tasks, including object classification~\cite{he2015residual}, instance segmentation~\cite{dai2015instance}, and 2D human pose estimation ~\cite{newell2016stacked}.

\section{Data Collection and Annotation}
With most modern machine learning frameworks, the data quantity and quality is as important, if not more so, than details of \changes{a} particular learning algorithm. While unlabeled data can be rapidly obtained, high quality labeled data is time consuming and costly to acquire. Recent efforts \cite{castrejon2017annotating,sohn2020fixmatch,xie2019unsupervised} leverage machine learning to assist with manual annotation or to provide semi-supervised labels from unlabeled data. Similarly, \cite{scenenn-3dv16} and \cite{marion2017labelfusion} leverage multi-view scene reconstructions to efficiently annotate 3D objects, then reproject the annotation masks to individual frames based on estimated camera viewpoints. \cite{osteen2019reducing} combined geometric reconstructions \changes{with hierarchical} metric segmentations \changes{as well as} learned object-segmentation proposals to annotate 3D scene objects, reprojecting label \changes{masks} to individual frames. Here, we extend \changes{this work} to add rapid keypoint annotation with depth-based keypoint refinement to enable downstream applications such as object pose estimation from color images.
The work is most closely related to \cite{marion2017labelfusion}, which also uses object reconstructions to annotate objects and their poses.
While their work 
\changes{requires}
3D meshes \changes{of the objects as a prerequisite to using the annotation tool}, we assume no existing models and emphasize the scenario of needing to rapidly annotate a completely unknown object. \changes{In addition, our tool specifically produces annotations for training keypoint detectors, \secondchanges{while} the alternative tool does not. As a result, we can leverage the structure of the object given by the 3D keypoint annotations to perform automatic keypoint refinement with no additional human work beyond keypoint labeling.}

To label a new object class with keypoint annotations, we collect sequences of \changes{streaming} data from different viewpoints,  environments, and lighting conditions using an RGB-D sensor. 
\changes{We annotate the resulting images by performing 3D reconstruction, manually annotating the resulting models, projecting the keypoints into the image frame, and refining the resulting keypoints}.

\changes{
\subsection{3D Reconstruction}
}
\label{sec:3d_recon}
We leverage state-of-the-art scene reconstruction from RGB-D sequences \cite{whelan2016elasticfusion} as well as surface generation algorithms \cite{kazhdan_13} to pre-process the collected data. \changes{In previous work \cite{osteen2019reducing}, we evaluated the performance of modern scene reconstruction algorithms, specifically Elastic Fusion \cite{whelan2016elasticfusion}, Elastic Reconstruction \cite{choi15elasticreconstruction}, and Kintinuous \cite{whelan2015kintinuous}.
}
\changes{
Kintinuous is an extension to the original KinectFusion reconstruction algorithm \cite{izadi2011kinectfusion} which supports operation over larger scales with loop closure. Kintinuous was designed for corridor-like motion with infrequent loop closures over large distances, rather than the potentially frequent loop-closures that can arise from the context of overlapping camera motion for object reconstruction \cite{whelan2016elasticfusion}.
Elastic Reconstruction is an offline reconstruction algorithm that constructs sets of scene fragments (or sub-models), each consisting of a small number (50) of integrated sequential RGB-D frames, and then globally registers the fragments and optimizes the associated sensor poses for a final reconstruction.
}

\changes{
In contrast to these algorithms, Elastic Fusion does not perform pose-graph optimization. Instead, Elastic Fusion is designed with a focus on the quality of the reconstructed model, formulating the problem using a deformation graph. A surfel map is used to model the environment, where surfels store properties such as color, normal, radius, etc. Frame-to-model sensor tracking is performed using the portion of the model that has been recently observed (\textit{active}), while global loop-closure candidates are determined by attempting to register new data with portions of the model that have not been recently observed (\textit{inactive}). During loop closure, the model is non-rigidly deformed to align surfels in the deformation graph, in contrast to pose-graph optimization approaches which do not perform non-rigid deformation of the data, instead optimizing the sensor trajectory given the measured data and associated uncertainty. After deforming sets of matching surfels according to surface correspondences, the camera poses are updated by applying the relative transform which brings the surfaces into alignment.
}

\changes{
The evaluation from~\cite{osteen2019reducing} on sequences from five classes of the Redwood dataset \cite{Choi2016} found that the reconstructions of Elastic Fusion were comparable in terms of accuracy to the offline Elastic Reconstruction
approach at a fraction of the runtime. Therefore, for this work we use Elastic Fusion to create scene models, though the annotation system is agnostic to the choice of reconstruction algorithm.
}

\changes{Once reconstruction has taken place, the resulting 3D model as well as the estimated camera poses from each frame are input to the annotation tool.}
The only phase of the process that requires human input is the placement of keypoints on the surface of the 3D reconstruction, after which keypoint annotations are automatically projected back to the original input image stream.

\subsection{Annotation Refinement}\label{sec:ann-refinement}

\changes{
An ideal reconstruction system would correctly estimate the camera poses for each frame of a sequence, such that the generated depth images match \secondchanges{corresponding} real depth images for all frames. In practice, pose-graph based SLAM techniques such as Elastic Reconstruction as well as deformation techniques such as Elastic Fusion attempt to minimize accumulated pose or model errors across the sequence. While this is optimal for the purposes of estimating sensor trajectory over a sequence, the global pose estimate for a given frame depends the accuracy of the estimates of other poses in the sequence. 
}

\changes{While we only annotate sequences for which qualitatively good reconstructions are produced, we observe
the presence of drift in the camera pose estimates over the course of a sequence. In many cases, this drift is mitigated by loop closure in the reconstruction algorithm, but we still develop techniques to account for potentially inaccurate camera pose estimates at any given frame.}

\changes{Specifically, we use the 3D model and camera trajectory to generate depth images at each camera pose, then perform data association between the generated depth images and the corresponding true depth images from the sensor. This technique ensures that the reconstruction results, integrated over the course of an entire sequence, are updated to align with the real data at each individual frame.
}
 While we could also compare input color images with their generated counterparts, we choose to focus on depth comparisons on the assumption that 3D reconstructions preserve object shape better than object appearance for any given viewpoint and lighting condition.
Therefore we use the depth images to refine annotations, \changes{and we compare two alternative approaches to refining annotations: those that update keypoints individually versus those that update all keypoints together.}

\changes{\subsubsection{Keypoint-level Refinement}}
\label{sec:feat-match}

\changes{The most noticeable artifact of keypoint projection with camera trajectory drift is pixels which project to the background rather than the object of interest. More subtly, visible keypoints may project to the correct object but be a few pixels away from the correct location, which may also be obvious to the observer depending on the choice of keypoint location.}

\changes{To address the issues of keypoints projecting to background objects ("jump-edge" pixels), as well as keypoints that are visible but not correctly located, we create an approach to refine individual keypoint locations. In order to determine if a keypoint is eligible for refinement, we first determine whether it is occluded or not for the current camera pose $\mathbf{T}^{F}_{C}$ in fixed frame $F$. Given $N$ keypoints specified for an object, the set of object keypoints is defined as \secondchanges{$K = \{{}^{F}\mathbf{k}_{0},{}^{F}\mathbf{k}_{1},..,{}^{F}\mathbf{k}_{N}\}$}.
}

\changes{
For each \secondchanges{${}^{C}\mathbf{k}_i \in \mathbb{R}^3$} in \secondchanges{$K$}, transformed from fixed frame $F$ to camera frame $C$ \secondchanges{by ${}^{C}\mathbf{k}_i=\mathbf{T}^{F}_{C} {}^{F}\mathbf{k}_i$}, the occlusion test is performed by comparing the depth of the predicted 3D keypoint location \secondchanges{${}^{C}\mathbf{k}_i$} with the depth value of the generated depth image at the pixel which the keypoint projects to, effectively performing a z-buffer test to identify which keypoints are occluded.
}
\changes{
Also, with knowledge of the 3D object geometry, local normals for each keypoint ${}^{C}\mathbf{n}_i$ are compared to the  \secondchanges{camera viewpoint ${}^{C}\mathbf{v}$, where ${}^{C}\mathbf{v} \in \mathbb{R}^3$ represents a unit vector along the principal axis of the camera}, and keypoints are identified as occluded if their local normals \secondchanges{are nearly orthogonal to or face away from} the camera. \secondchanges{More specifically, a keypoint is determined to be occluded if ${}^{C}\mathbf{v} \cdot {}^{C}\mathbf{n}_i > \tau$, where $\tau$ is an approximate orthogonality threshold that is set to -0.15 radians for this evaluation.} }


    
    

\changes{
The following keypoint-level refinement methods are mutually exclusive, such that no two refinements will be applied to the same keypoint.
}

\textbf{Jump-edge adjustment}: \changes{Often, corners and edges of objects are good candidates for keypoint locations. For even small camera pose estimation errors, the projected keypoint location can jump past the object of interest to the background. This is addressed by an adjustment approach that takes advantage of the fact that our 3D keypoint positions are known for each estimated camera pose, and that we can easily compare the depth values of the \secondchanges{generated depth image and real depth image} at the keypoint's projected pixel location to identify jump-edges. Specifically, we use our reconstructed model to project the keypoints into the estimated camera frame. Any keypoint where the projected depth is less than the measured depth at its projected pixel coordinates is deemed a jump edge. If a jump edge is detected, the refined keypoint location is updated to be the projection of the closest point in the real depth image to the estimated 3D keypoint location.}

\textbf{Feature matching}: \changes{For keypoints that correctly project to the desired object, 3D feature matching is employed to improve the keypoint location estimate. If a keypoint is clearly visible for the given camera pose, then a 3D descriptor is extracted from the generated depth image and compared to descriptors extracted in the true depth image around the estimated keypoint location. For this study, we use the FPFH feature descriptor \cite{FPFH}, \changes{which was shown by \cite{choi15elasticreconstruction} to be highly effective at geometric pairwise registration}, but have also tested with the SHOT descriptor \cite{Salti2014SHOTUS}. The updated keypoint is defined by a weighted sum of the Euclidean distance and the matched feature vector distances\secondchanges {(with weight values of 0.7 and 0.3, respectively)} between the real and generated depth images. The closest match in the real depth image is then chosen as the refined keypoint location.
}

    



    



\changes{
\subsubsection{Object-level Refinement}\label{sec:icp}
}

\changes{
Since the goal of annotation refinement is to maximize the alignment of the annotated model and an individual frame, regardless of any other camera pose estimates in the trajectory, we perform iterative closest point (ICP) on the generated and real depth images to ensure the model is best aligned with each depth frame. For this work, our primary concern is to align points belonging to the object of interest, so we limit our ICP refinement to the bounding volumes that contain the annotated objects. We use the volume that minimally bounds the annotated keypoints to define the volume in the generated image, and since there may be error in the pose estimate, the volume is dilated for the real depth image. Then, point-to-plane ICP \cite{chen1992ICP} is performed on the cropped points from the generated and real point clouds derived from the respective depth images. The camera pose estimate is adjusted based on the ICP alignment, and all keypoints are transformed accordingly. This means that all keypoints undergo the same rigid transformation, which preserves the relative poses of the keypoints, and allows us to refine the positions of \secondchanges{both} visible and occluded keypoints.
}

\changes{
\section{\secondchanges{RCTA Object Keypoints Dataset}}
\label{sec:dataset}
While large scale datasets such as \secondchanges{PASCAL3D+}  contain many types of objects, there are \secondchanges{still} many objects and environments for which there is no corresponding dataset.
This is especially true for the \secondchanges{Robotics Collaborative Technology Alliance (RCTA)}  program, which is a consortium of government, industry, and academic researchers focusing on fielding embodied autonomous systems in challenging and unstructured environments. The scope of the work includes many unique objects that are not well-represented in existing datasets, such as the Czech Hedgehog.

As part of this work, we have therefore created a dataset containing 101 sequences split across six of the objects used by the RCTA~\cite{spie_perception} in both indoor and outdoor environments: \textit{barrel}, \textit{barrier}, \textit{crate}, \textit{gas can}, \textit{Czech Hedgehog}, and \textit{robot}. \secondchanges{The number of keypoints for each object class are shown in Table~\ref{tab:rcta_keypoint_numbers}.}

We collected many data sequences for each object, not knowing ahead of time how many would yield accurate 3D reconstructions. Indeed, many reconstructions did fail or were not accurate enough to include in the analysis. To avoid adding the burden to a user of filtering poor reconstructions manually, we defer the determination of reconstruction viability to the actual annotation process and allow the annotator to trivially skip a poor reconstruction. In total, we collected \secondchanges{283 sequences across the six object classes, and determined that 101} were suitable for annotation, some of which are shown in Figure~\ref{fig:example_reconstructions}. For each object class, a random subset of sequences and frames were annotated by hand. The total number of frames analyzed for this work is 10281; of those, 951 were manually annotated.

With the public release of this dataset, we hope to increase diversity in the types of environments and objects used for keypoint annotation. Furthermore, we release the raw data collection sequences as well as the reconstructed models, with the hope of testing new reconstruction algorithms as they are released.

\secondchanges{The full dataset is available for download at \url{https://sites.google.com/view/rcta-object-keypoints-dataset}.} 

\begin{table}[]
    \centering
    \begin{tabular}{c|c|c|c|c|c|c}
        \secondchanges{Class} & Barrel & Barrier & Crate & Gas Can & Hedgehog & Robot\\
         \hline
        \secondchanges{Number of Keypoints} & 6 & 9 & 12 & 10 & 6 & 11 \\
    \end{tabular}
    \caption{\secondchanges{Number of keypoints for each class in the RCTA Object Keypoints Dataset}}
    \label{tab:rcta_keypoint_numbers}
\end{table}

\begin{figure}
\begin{subfigure}{0.49\columnwidth}
    \includegraphics[width=\linewidth]{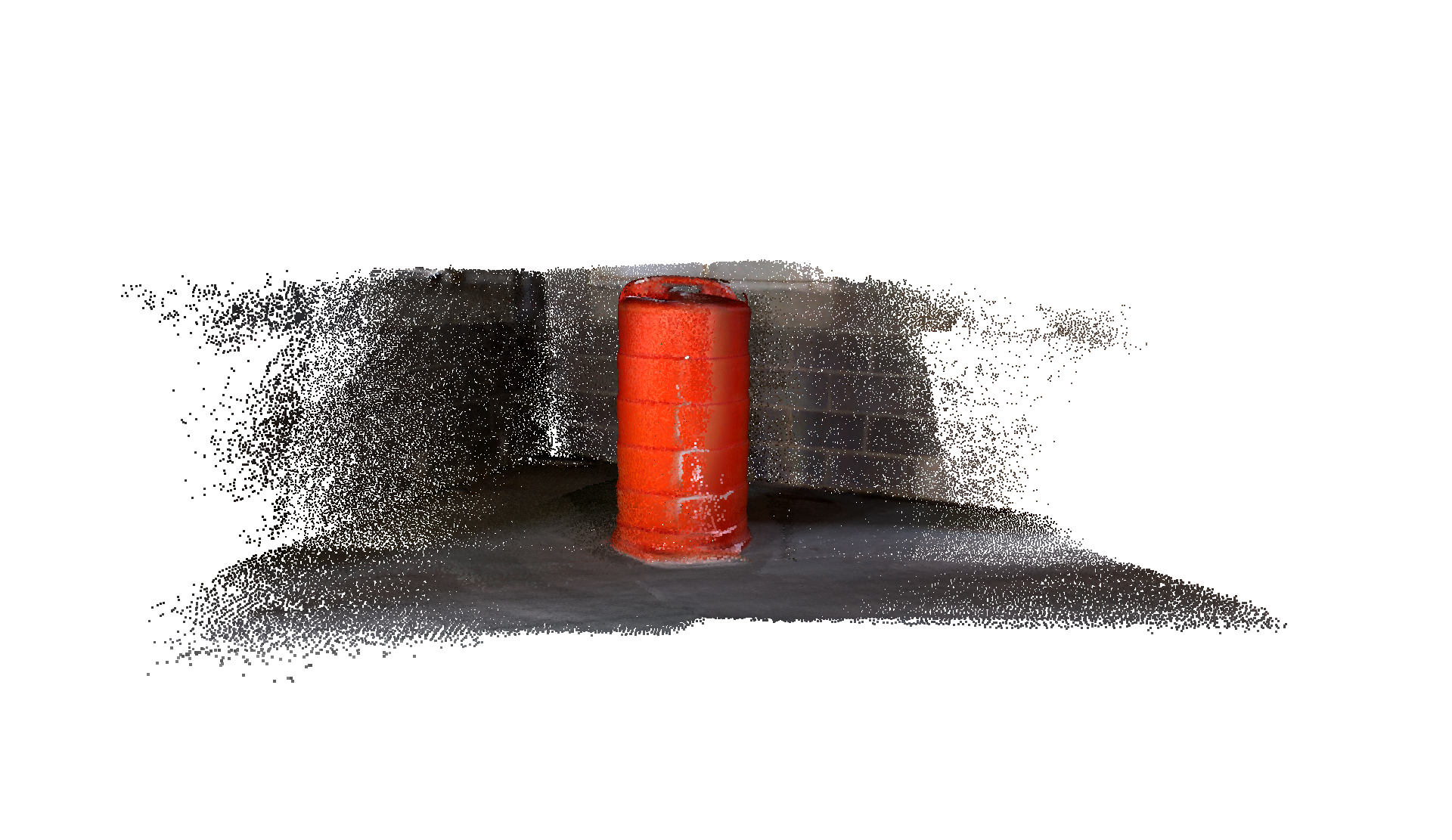}
    \caption{Barrel}
\end{subfigure}\hfill
\begin{subfigure}{0.49\columnwidth}
    \includegraphics[width=\linewidth]{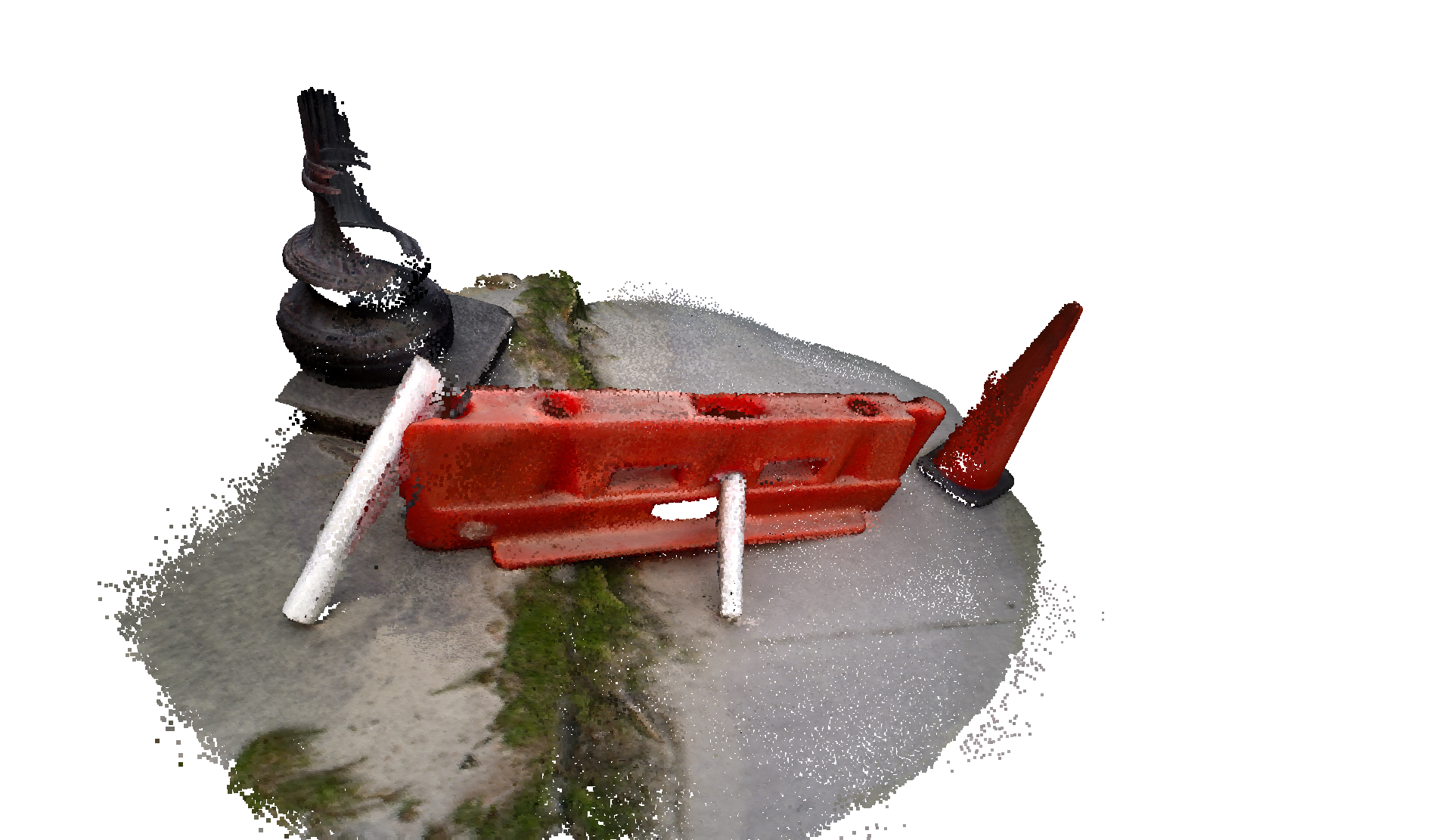}
    \caption{Barrier}
\end{subfigure}\hfill
\begin{subfigure}{0.49\columnwidth}
    \includegraphics[width=\linewidth]{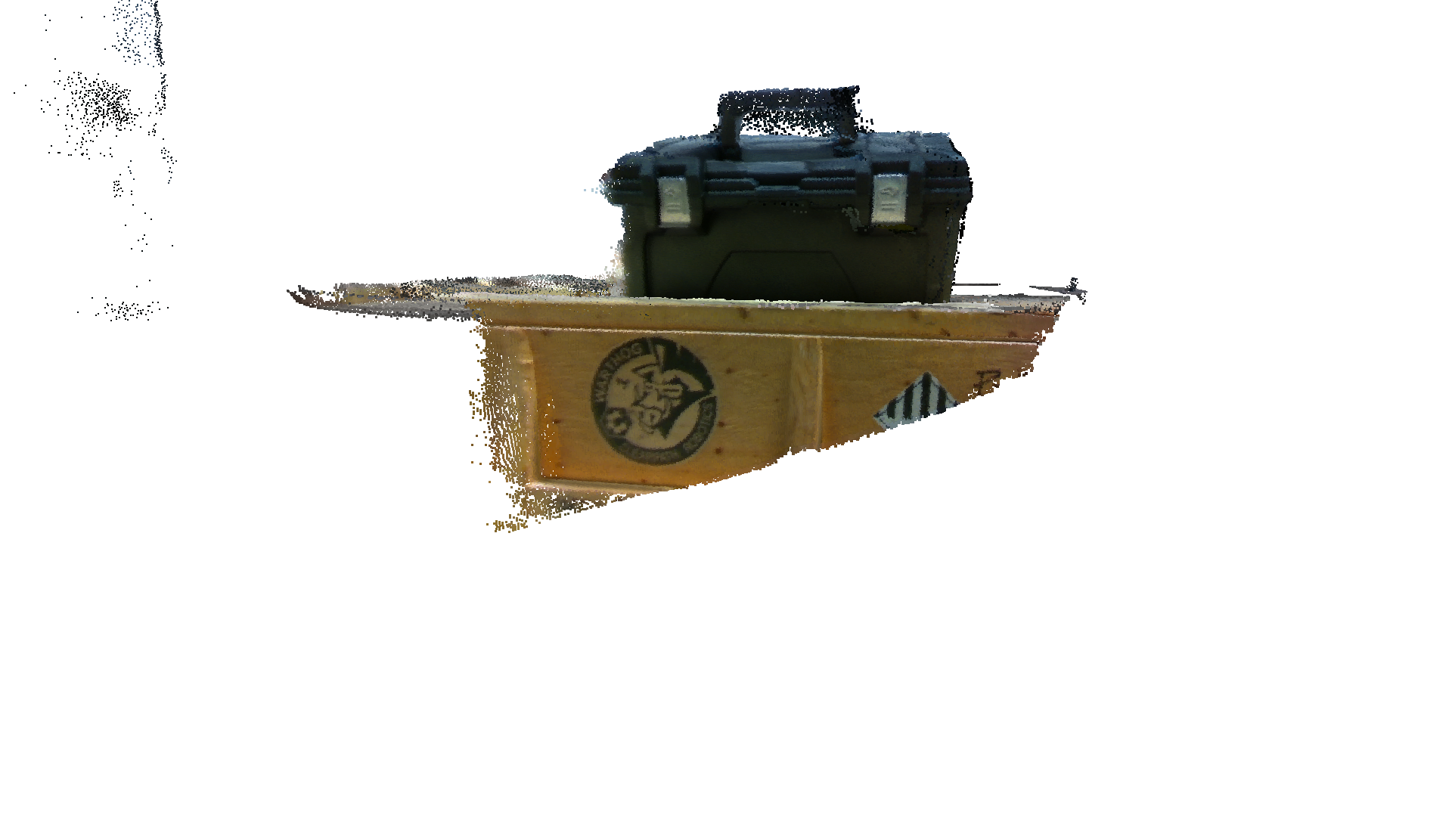}
    \caption{Crate}
\end{subfigure}\hfill
\begin{subfigure}{0.49\columnwidth}
    \includegraphics[width=\linewidth]{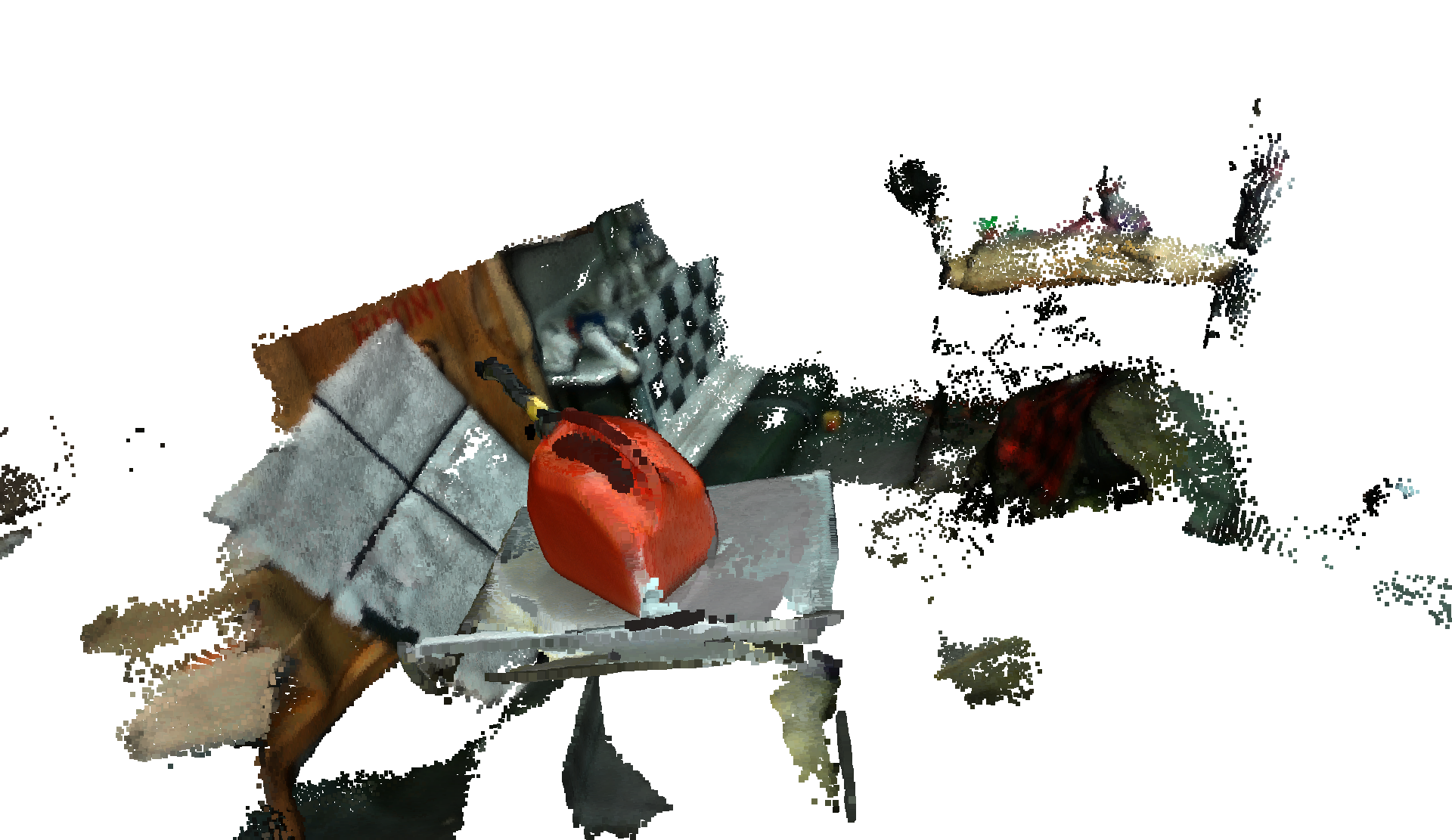}
    \caption{Gas Can}
\end{subfigure}\hfill
\begin{subfigure}{0.49\columnwidth}
    \includegraphics[width=\linewidth]{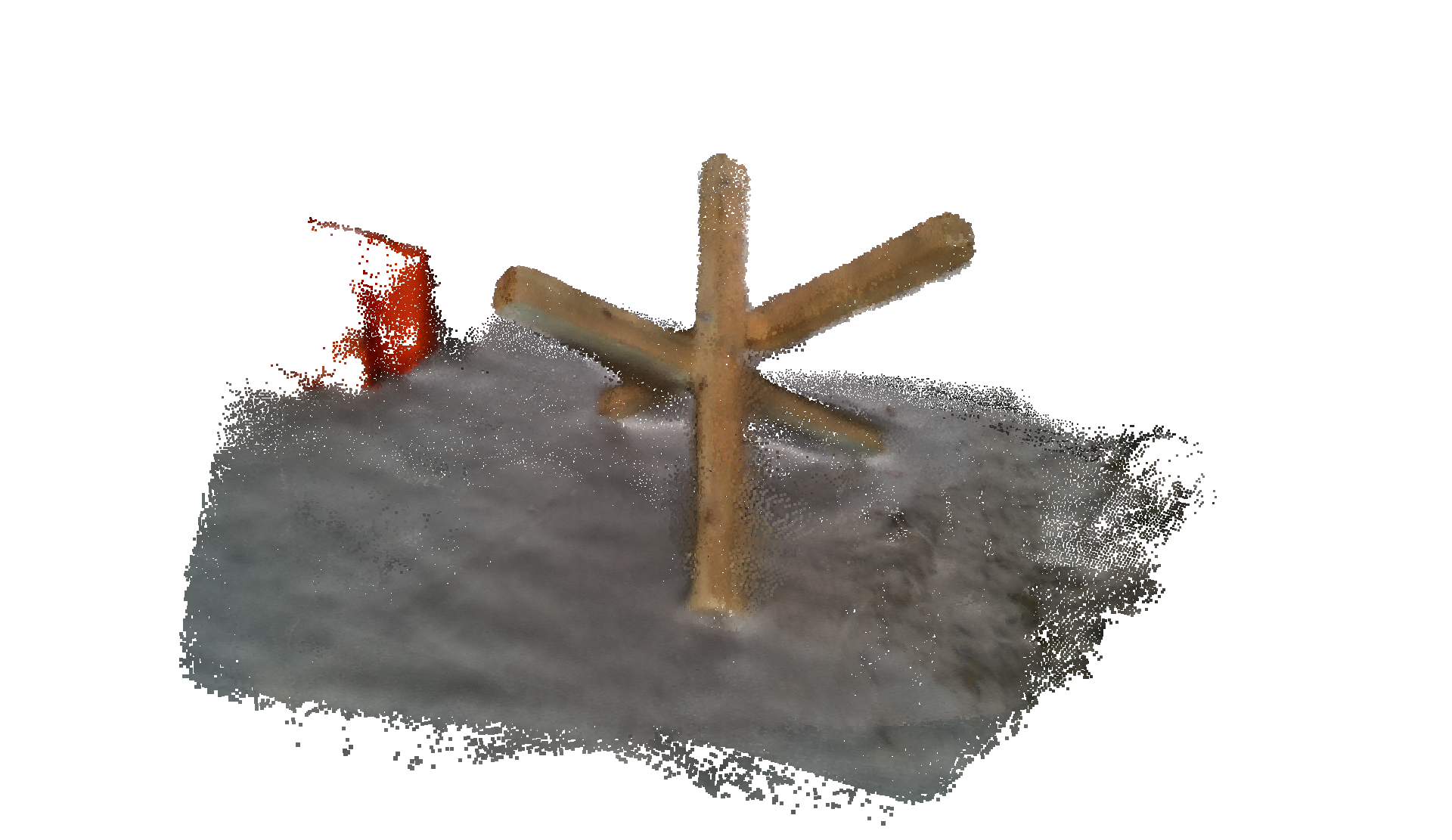}
    \caption{Hedgehog}
\end{subfigure}\hfill
\begin{subfigure}{0.49\columnwidth}
    \includegraphics[width=\linewidth]{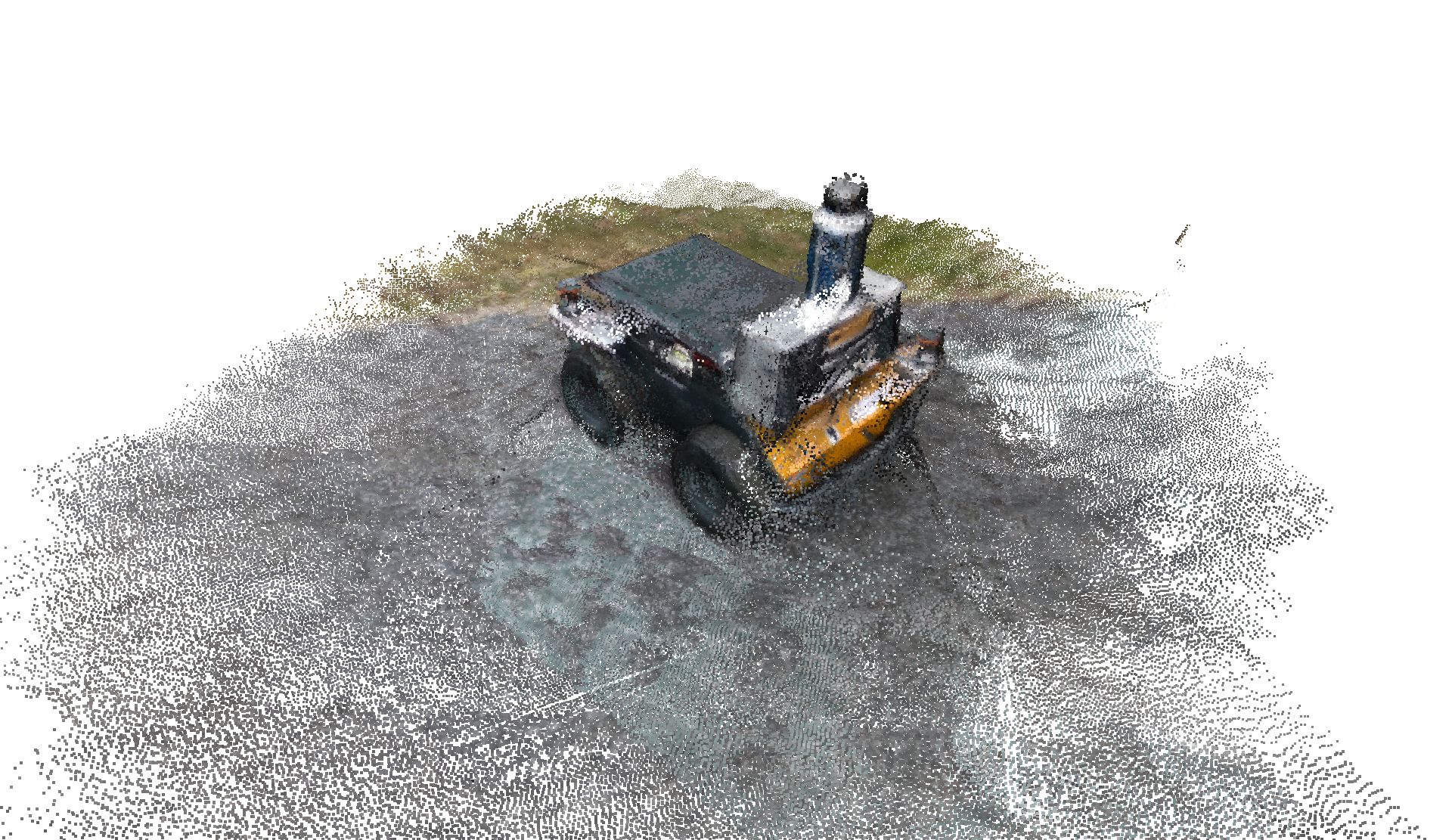}
    \caption{Robot}
\end{subfigure}\hfill
    \caption{Example 3D reconstructions of the objects in the RCTA dataset.}
    \label{fig:example_reconstructions}
\end{figure}
}

\section{Results}

\thirdchanges{We present an analysis of our} architecture on existing benchmark datasets as well as custom data collections. In both qualitative and quantitative output, we show the effectiveness of the approach, and demonstrate the challenges of acquiring accurate manual annotations for occluded keypoints.

\subsection{Class-based pose recovery: PASCAL3D+}\label{sec:pascal}
We demonstrate the full strength of our approach using the large-scale PASCAL3D+ dataset \cite{xiang2014}. The stacked hourglass network was trained from scratch with the training set of PASCAL3D+. Instead of training separate models for different object classes, a single network was trained to output heatmap predictions for all of the 88 keypoints from all classes. \secondchanges{The number of keypoints for each class are shown in Table~\ref{tab:pascal_keypoint_numbers}.} Using a single network for all keypoints allows us to share features across the available classes and significantly decreases the number of parameters needed for the network. At test time, given the class of the test object, the heatmaps corresponding to the keypoints belonging to this class were extracted. For pose optimization, two cases were tested: (i) the CAD model for the test image was known; and (ii) the CAD model was unknown and the pose was estimated with a deformable model whose basis was learned by PCA on all CAD models for each class in the dataset. Two principal components were used ($k=2$) for each class, which was sufficient to explain greater than $95\%$ of the shape variation. The 3D model was fit to the 2D keypoints with a weak-perspective model, as the camera intrinsic parameters were not available.

\begin{table}[]
    \centering
    \begin{tabular}{c|c|c|c|c|c|c|c|c|c|c}
        \secondchanges{Class} & aero & bike & bottle & bus & car & chair & sofa & train & TV monitor & boat\\
         \hline
        \secondchanges{Number of Keypoints} & 8 & 11 & 7 & 12 & 12 & 10 & 10 & 8 & 4 & 6 \\
    \end{tabular}
    \caption{\secondchanges{Number of keypoints for each class in PASCAL3D+}}
    \label{tab:pascal_keypoint_numbers}
\end{table}

\begin{figure*}
  \centering
  \includegraphics[width=\linewidth]{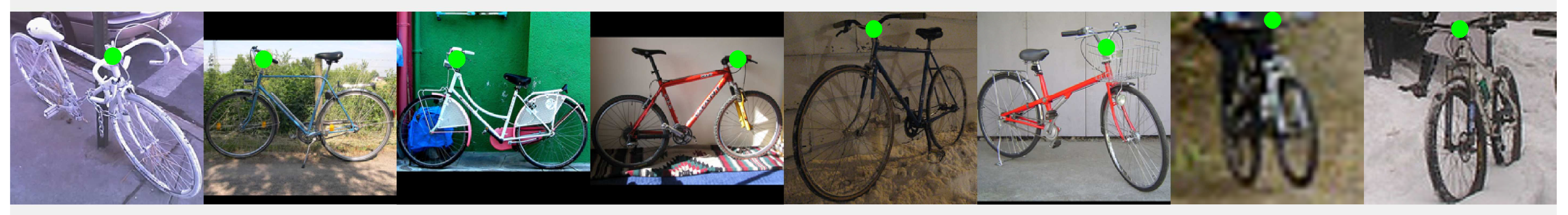}\\
  \includegraphics[width=\linewidth]{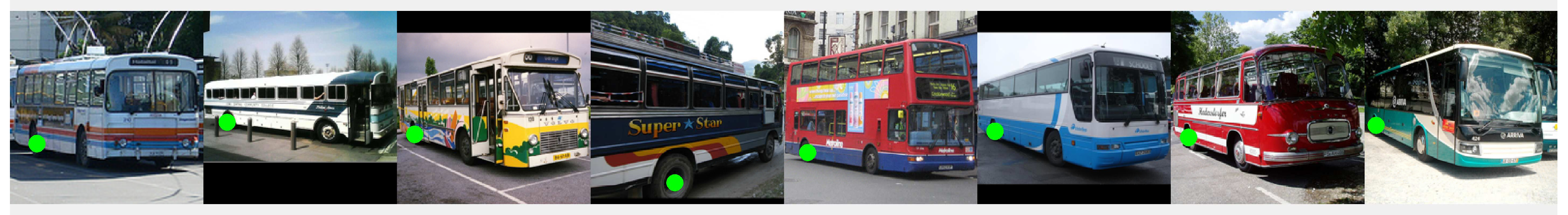}\\
  \includegraphics[width=\linewidth]{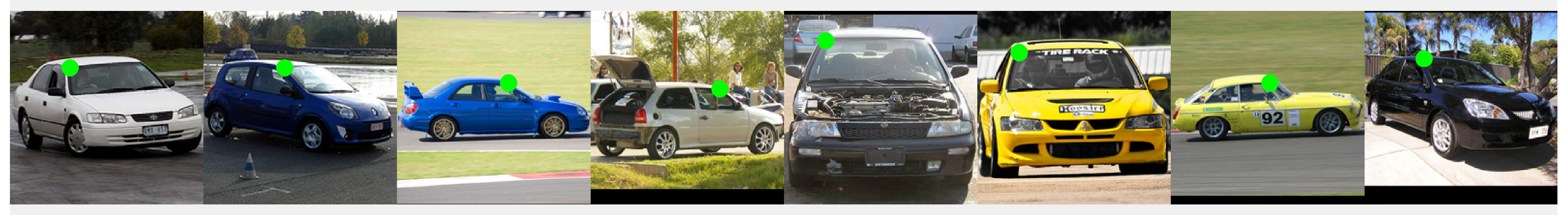}\\
  \includegraphics[width=\linewidth]{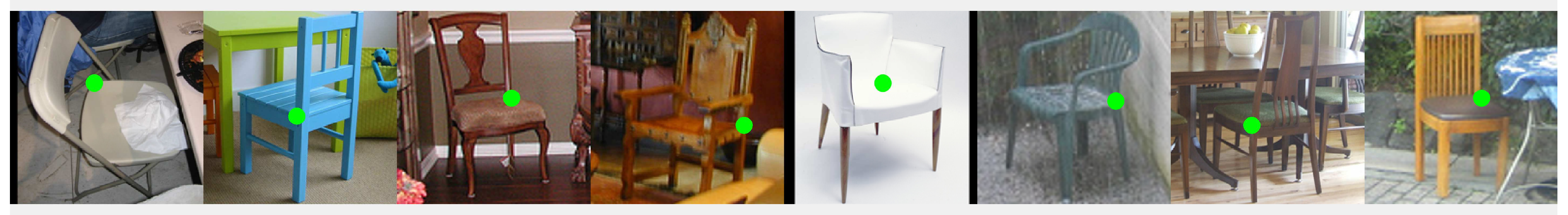}\\
  \caption{Localization results for diverse keypoint categories. We visualize eight images per category in each row selected randomly from the top 50 responses for each keypoint. The keypoint localization network is particularly successful at establishing semantic correspondences across the instances of a class, despite the significant intra-class variation and wide ranging camera viewpoints.}\label{fig:semantic}
\end{figure*}

\vspace{5pt}\noindent{\bf Semantic correspondences } A crucial component of our approach is the powerful learning procedure that is particularly successful at establishing correspondences \secondchanges{between the semantically related keypoints across different instances of an object class}. To demonstrate this network property,  in Figure~\ref{fig:semantic} we present a subset of the keypoints for each class, along with the localizations of these keypoints, in a randomly-selected set of images among \thirdchanges{those} with the top 50 responses. It is interesting to note that, despite the large appearance differences due to extreme viewpoint and intra-class variability, the predictions are very consistent and preserve the semantic relation across various class instances.

\vspace{5pt}\noindent{\bf Pose estimation } The quantitative evaluation for pose estimation on PASCAL3D+ is presented in Table~\ref{tab:pascal3d}. 
\thirdchanges{We only report rotation errors, as }
the 3D translation cannot be determined in the weak perspective case
\thirdchanges{nor is the ground truth available.}
\changes{Following work from \cite{tulsiani2015vk}, we use the geodesic distance in Equation~\ref{eq:geodesic} to calculate the rotational error between a pose estimate, $\bf{R_1}$, and the groundtruth, $\bf{R_2}$.}

\begin{align}
\secondchanges{\Delta(\bf{R_1},\bf{R_2}) = \frac{\|\log(\bf{R_1}^T\bf{R_2})\|_{F}}{\sqrt{2}}.}\label{eq:geodesic}
\end{align}

\begin{figure*}
  \centering
  \includegraphics[width=0.49\linewidth]{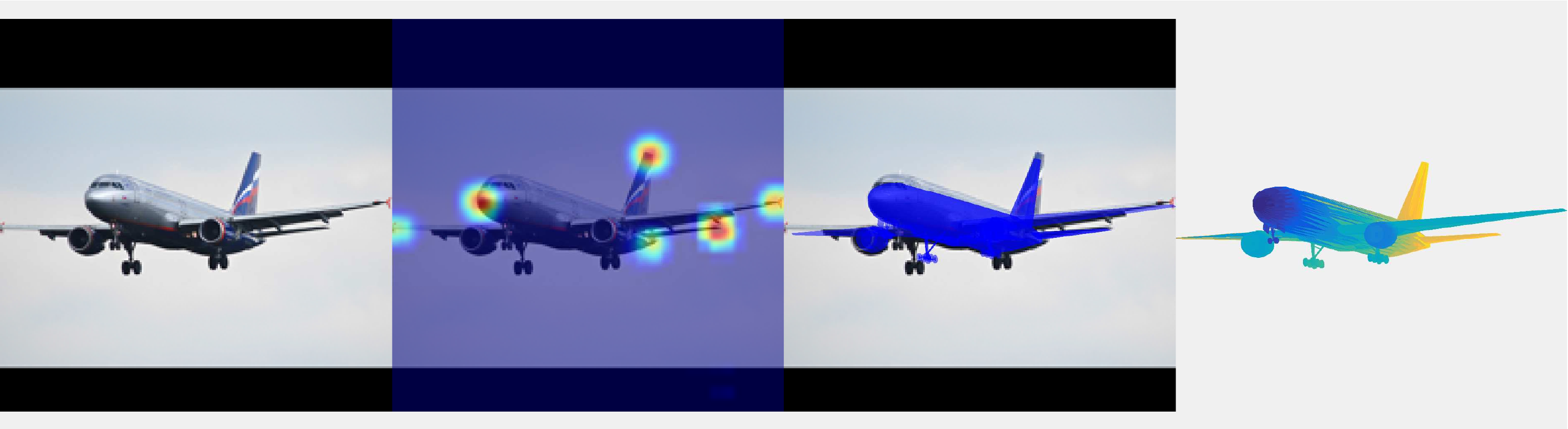}
  \includegraphics[width=0.49\linewidth]{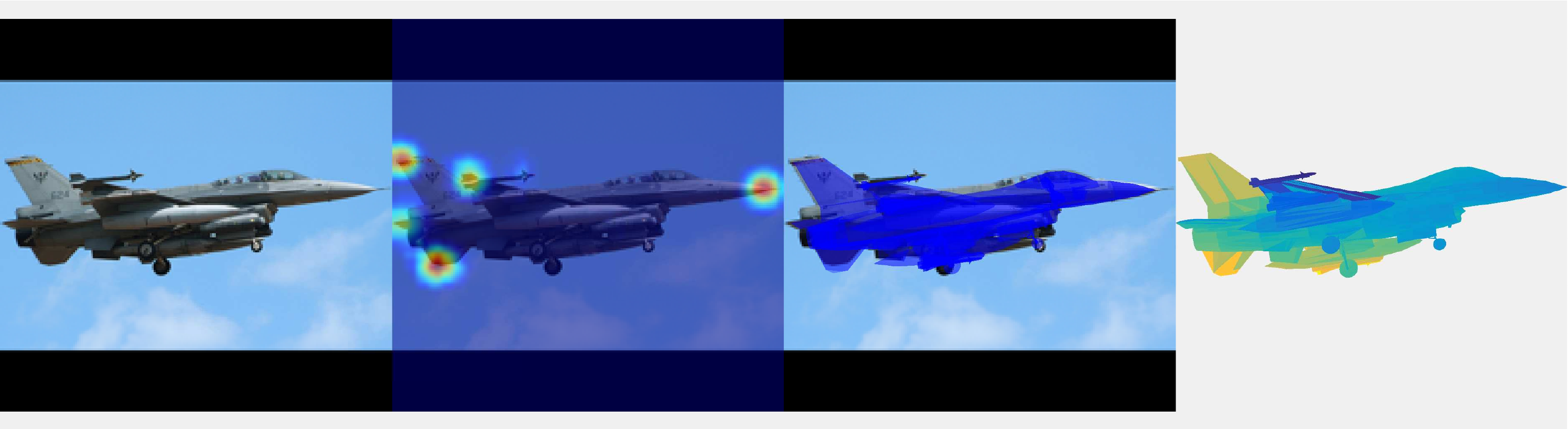}
  \includegraphics[width=0.49\linewidth]{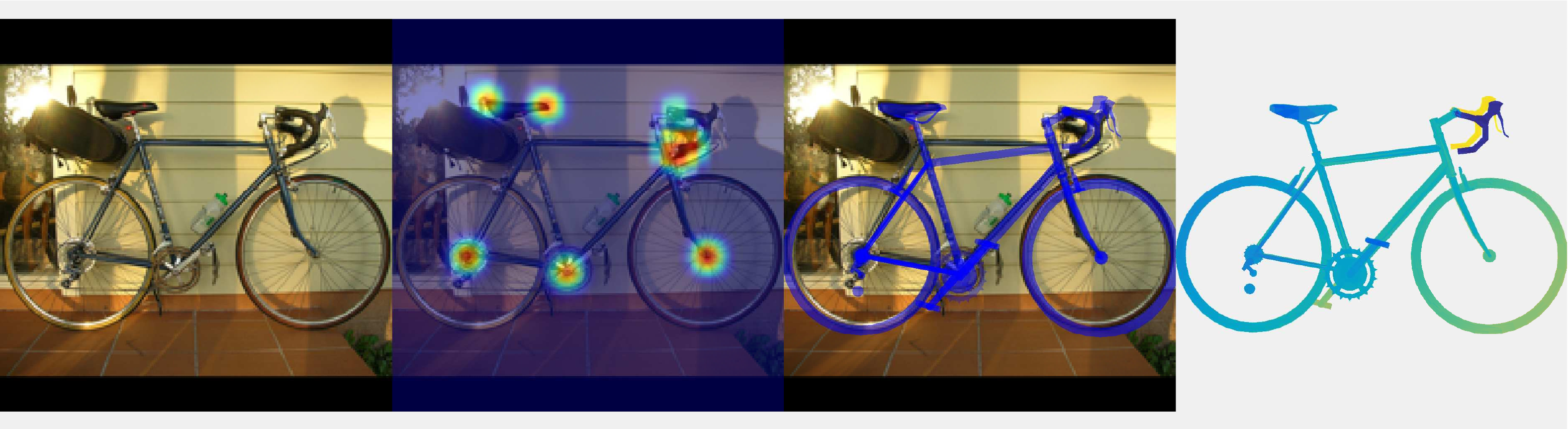}
  \includegraphics[width=0.49\linewidth]{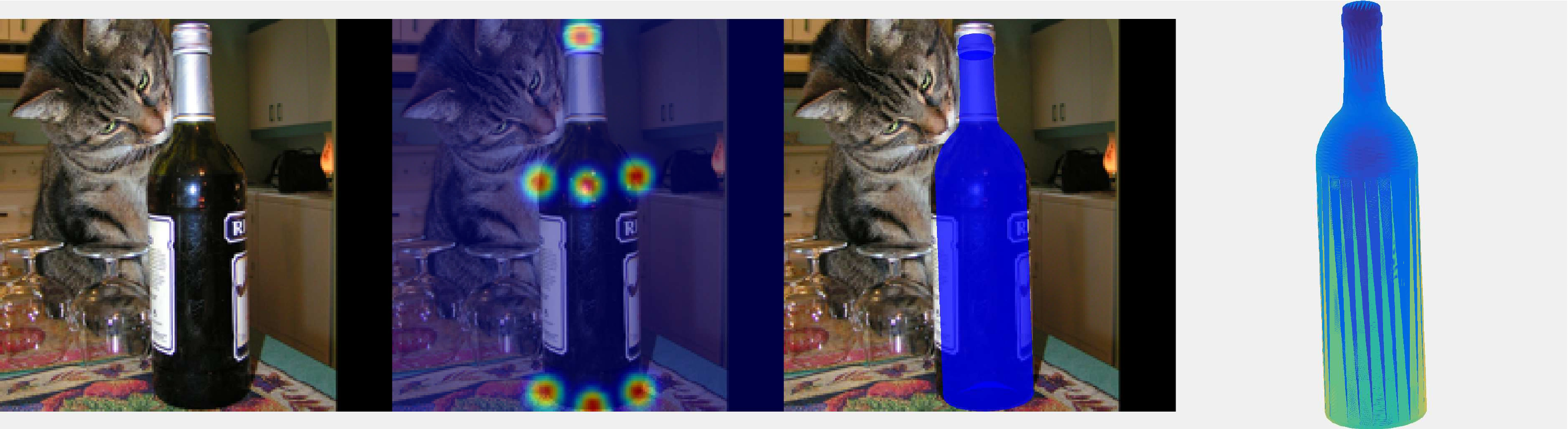}
  \includegraphics[width=0.49\linewidth]{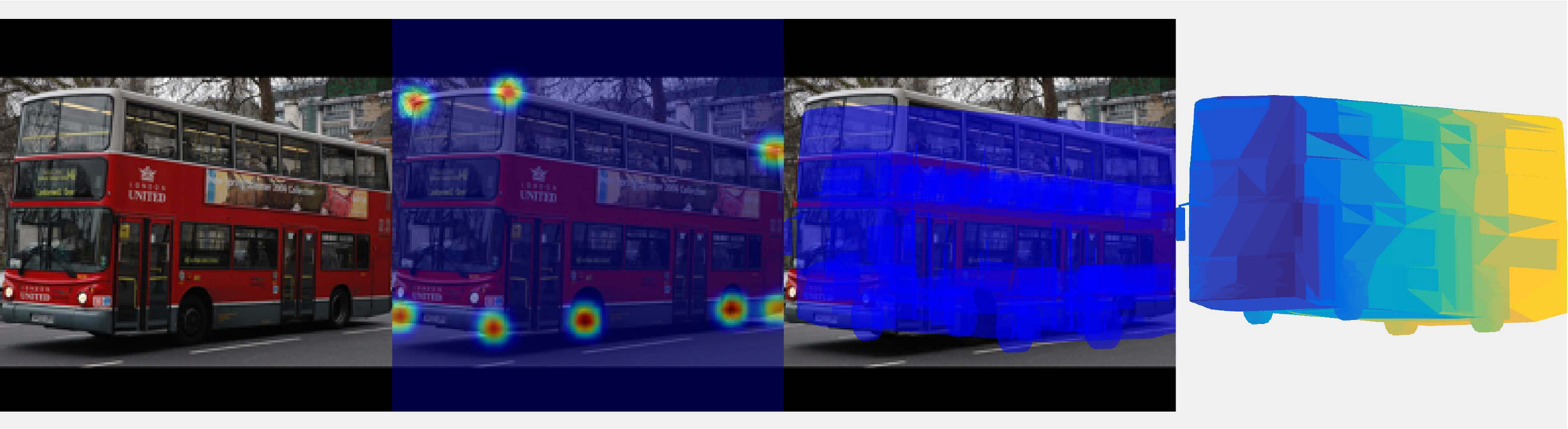}
  \includegraphics[width=0.49\linewidth]{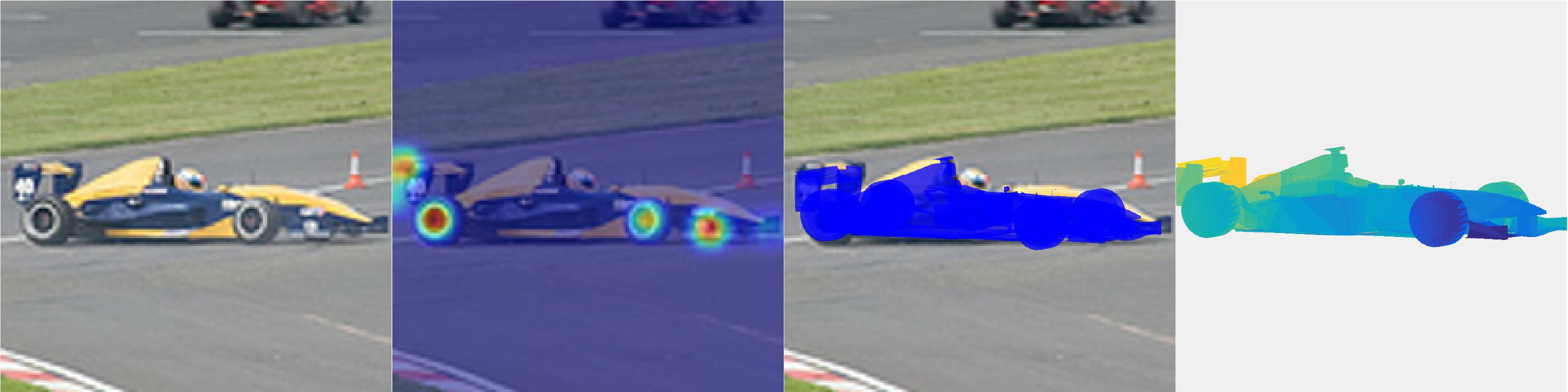}
  \includegraphics[width=0.49\linewidth]{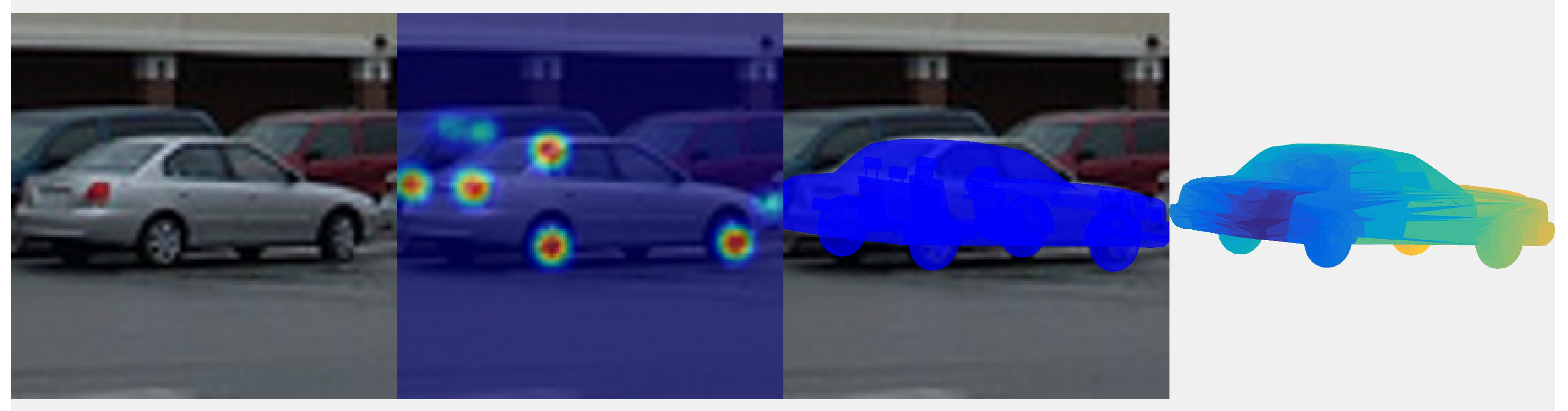}
  \includegraphics[width=0.49\linewidth]{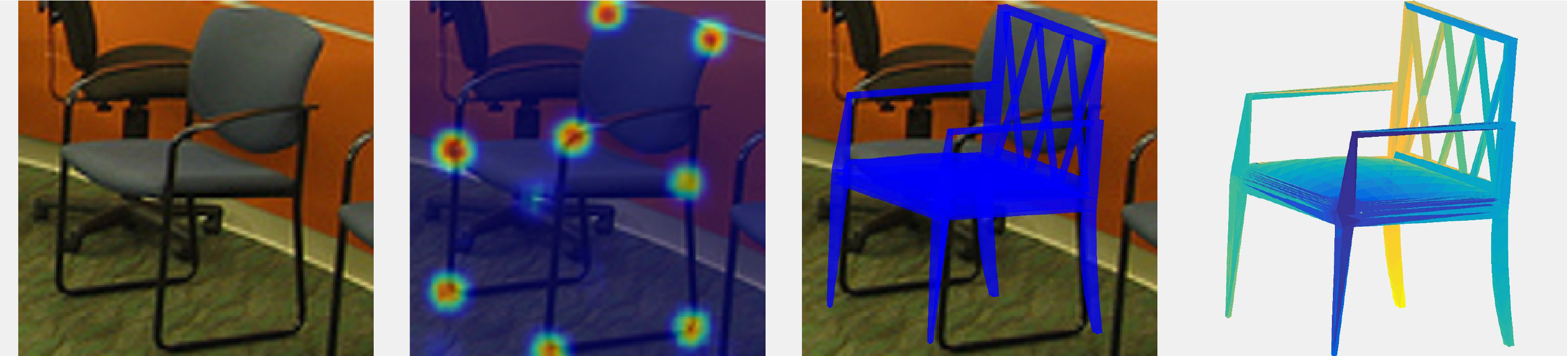}
  \includegraphics[width=0.49\linewidth]{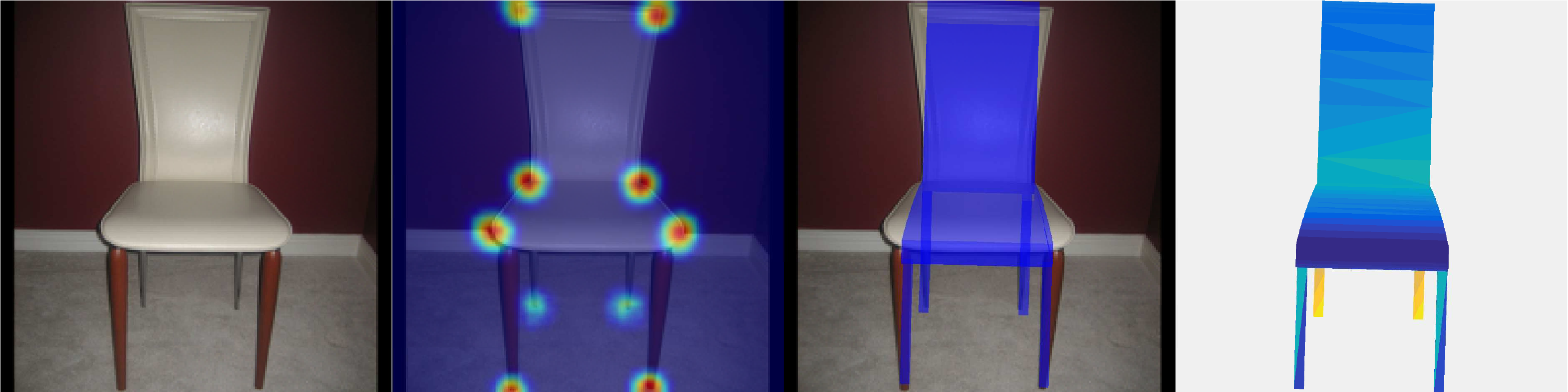} 
  \includegraphics[width=0.49\linewidth]{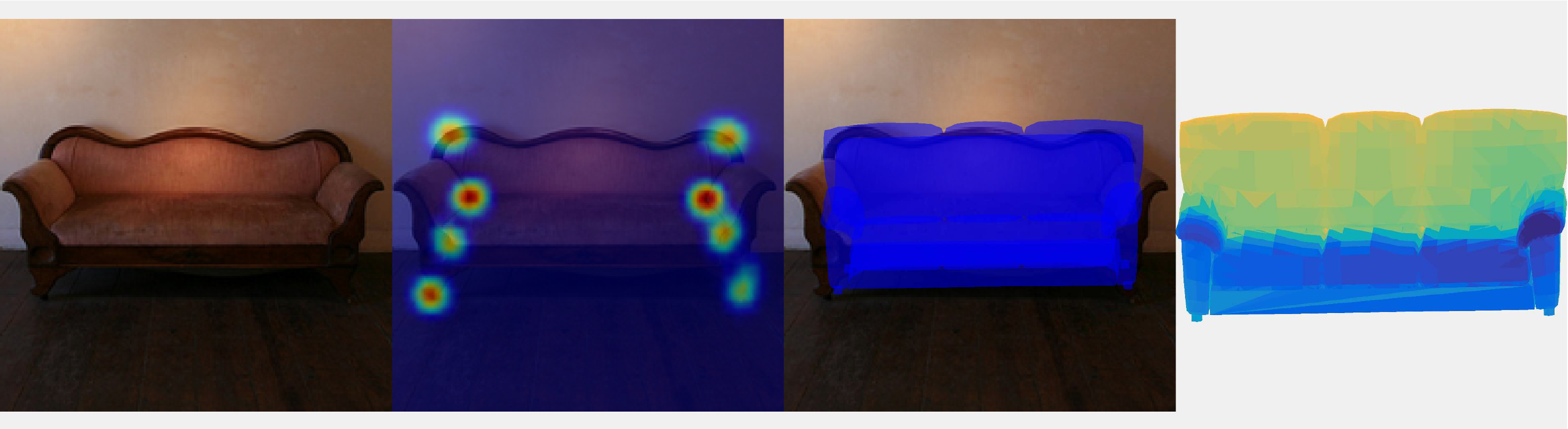} 
  \includegraphics[width=0.49\linewidth]{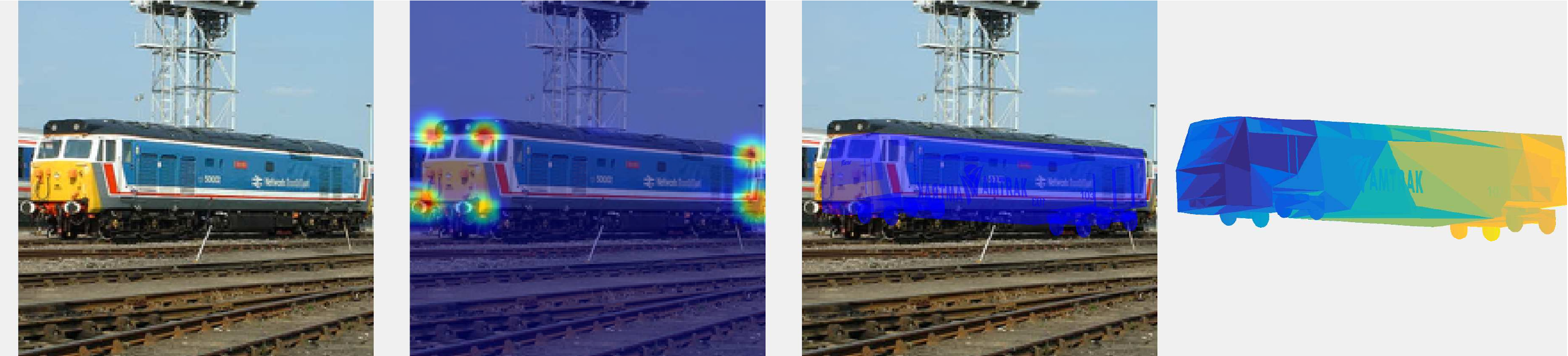}
  \includegraphics[width=0.49\linewidth]{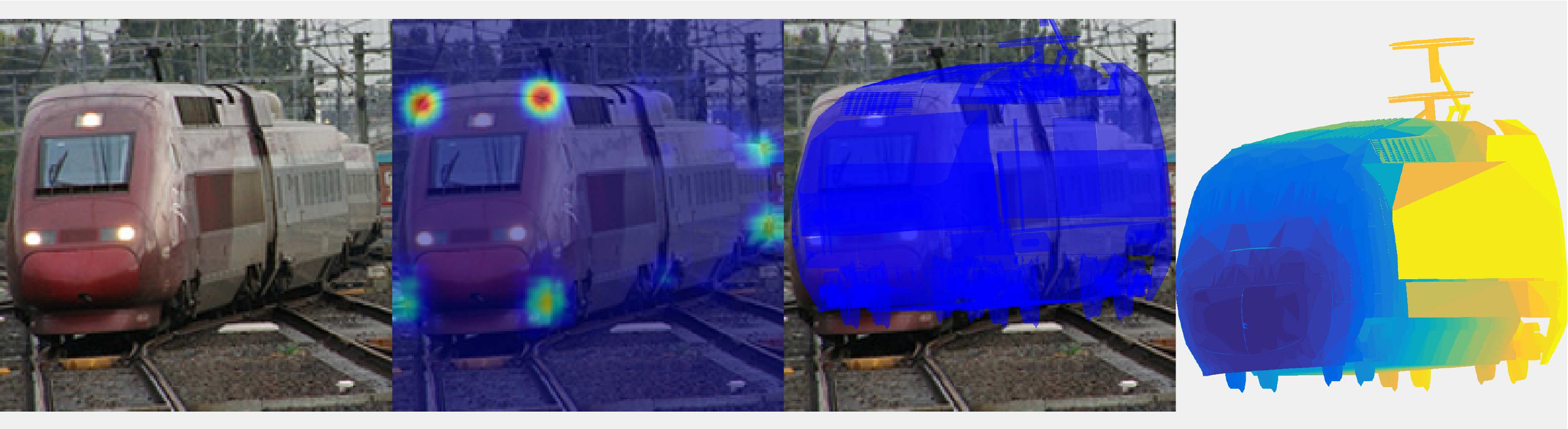}
  \caption{Example results of our approach on PASCAL3D+. For each example from left-to-right: 
the RGB image of the object, heatmap responses for the keypoints of the specific class, the CAD model projected to 2D after pose estimation, and the CAD model visualized in 3D.
}\label{fig:models3d}
\end{figure*}

\changes{where $\|x\|_{F}$ denotes the Frobenius norm of $x$.}
\thirdchanges{Our} method shows improvement across \secondchanges{several} categories with respect to the state-of-the-art. The best results are achieved in the case where the instance subclass for the object is known and there exists an accurate CAD model correspondence. 
\thirdchanges{Our} method with uniform weights for all keypoints is also compared as a baseline, which is worse than considering the confidences during model fitting.
\secondchanges{The worst results occur on object classes that were labeled with few semantic keypoints per instance.  The numbers of keypoints for each class in the dataset are shown in Table~\ref{tab:pascal_keypoint_numbers}. The two classes which performed worst for our method, the TV Monitor and the boat, had four and six keypoints respectively, while the remaining classes averaged just under ten keypoints each.}
 A subset of results of 
 \thirdchanges{our} method are visualized in Figure~\ref{fig:models3d}.
\thirdchanges{While our approach applies to both instance- and category-level pose estimation, it yields comparable results to other leading, but more limited methods.}

\begin{table*}[t]
\begin{center}
\begin{adjustbox}{width=1\textwidth}
\begin{tabular}{c|cccccccccc}
\hline
   Approach    &     aero &       bike &  	 bottle &           bus &           car &         chair &	sofa &         train &         TV monitor  &	boat        \\
\hline
Tulsiani and Malik~\cite{tulsiani2015vk}	&         13.8 &         17.7 &     	12.9 &          5.8 &          9.1 &         14.8 &         15.2 &          8.7 &         15.4 &    	{\bf 21.3}  \\
\changes{Deep3DBox~\cite{mousavian20173d}} & 13.6 & {\bf 12.5} & {\bf 8.3} & 3.1 & 5.8  & 11.9 & 12.8 & 6.3 & 11.9 & 22.8 \\
\changes{RenderForCNN~\cite{su2015render}} & 15.4 & 14.8 & 9.3 & 3.6 & 6.0 & {\bf 9.7} & {\bf 9.5} & {\bf 6.1} & {\bf 12.6} & 25.6\\
ours - PCA basis  &	11.2 &         15.2 &          13.1 &          4.7 &          6.9 &         12.7 &        21.7 &          9.1 &        38.5 &         37.9   \\
ours - CAD basis 	&   {\bf 8.0} &         13.4 &         11.7 &          {\bf 2.0} &          {\bf 5.5} &         10.4 &          9.6 &          8.3 &          32.9 &         40.7 \\
ours - uniform weights 	&	16.3 &         17.8 &          14.1 &          11.7 &          30.7 &         17.6 &        32.4 &          20.8 &        25.0 &         72.0   \\
\hline
\end{tabular}
\end{adjustbox}
\end{center}
\caption{Viewpoint estimation median error (degrees) on PASCAL3D+.}
\label{tab:pascal3d}
\end{table*}



\changes{Category-level pose is useful for robotic applications. In \cite{spie_t3}, the authors demonstrated a system for mobile manipulation, wherein our keypoint-based pose estimation algorithm was used to position the mobile manipulator relative to the object.  \thirdchanges{First, the mobile manipulator approached the target object and oriented itself to grasp an object of that object class.  Then a second} algorithm used a short-range depth sensor to select grasp locations from the generated pointcloud.  This approach is applicable, despite variation within an object class, as long as the variation between grasp locations in a class is smaller than the size of the robot's reachable workspace. }

\changes{Category-level pose estimation has also been used for semantic SLAM \cite{bowman2017probabilistic}, \thirdchanges{in which application the ability} to localize a wide variety of objects within a class is critical to populating the generated map with sufficient density.  \secondchanges{In the context of semantic SLAM, }being able to \thirdchanges{locate the surface of an object precisely} is less important than being able to estimate consistent and accurate poses for every encountered object of the target classes.}
\thirdchanges{Thus keypoint-based methods that provide category level pose can perform well while methods that densely reconstruct the surface of an object are not required.}

\changes{\subsection{Instance-based pose recovery: Benchmark for \secondchanges{6-DoF} Object Pose Estimation}
\label{sec:bop}

Our method is equally applicable in estimating object-level pose when the 3D object model and the camera intrinsic are known. This is the classic \secondchanges{6-DoF} pose task, and fits well with many robotics applications, where recognizing \secondchanges{6-DoF} pose allows better interactions with a known object. We demonstrate the effectiveness of our method on this task by comparing its performance to state-of-the-art methods on three object pose datasets.

To apply our method, we first set \secondchanges{${\bfc_i}={\bf0}$} \thirdchanges{for all $i$ in Equation~\ref{eq:cost}}. We derive \secondchanges{${\bfB}_0$} by choosing recognizable landmarks on the 3D model. \secondchanges{For symmetric objects as defined by the BOP benchmark, we place landmarks on geometries or surface texture that break the symmetries. As a result, \thirdchanges{our pose solution is more accurate than the minimum required}. No additional step is taken to handle symmetries.} For each training image having a ground truth, \secondchanges{6-DoF} pose, we reproject these 3D landmarks to generate 2D keypoints, from which the keypoint detector can be trained. The optimization procedure is the same as in category-level estimation but with \secondchanges{${\bfc}$} fixed at zero. Additionally, we fine-tune a Faster-RCNN~\cite{ren2015faster} object detector for each dataset to provide \thirdchanges{cropping bounding boxes} for our \secondchanges{method} following standard practice~\cite{hodan2020bop}.

\thirdchanges{The three object-pose datasets we use are}
LineMOD-Occluded~\cite{brachmann2014learning}, YCB-Video~\cite{calli2015ycb} and TUD-Light~\cite{hodan2018bop}. More specifically, we use the train-test split provided by the new BOP benchmark on these three datasets~\cite{hodan2020bop}. Using a standard split allows us to compare our method with other state-of-the-art methods while \secondchanges{minimizing the} performance difference that stems from using different data-generation techniques. The training set comprises synthetic data from physically-based rendering, and real data is added when it is available for that dataset.

\textbf{Evaluation metrics}. \thirdchanges{On each dataset we} evaluate our method with the three metrics used in the BOP19/20 challenge.
\begin{itemize}
    \item Visible Surface Discrepancy (VSD) calculates the average 3D distance between the estimated and the ground-truth objects on the visible part of the object
    \item Maximum Symmetry-Aware Surface Distance (MSSD) computes the maximum 3D distance between the estimated and the ground-truth objects. MSSD is similar to the classical \secondchanges{6-DoF} pose metric used to compute average distance, but Hodan \secondchanges{et al.}~\cite{hodan2018bop} argue that MSSD is less sensitive to object geometry
    \item Maximum Symmetry-Aware Projection Distance (MSPD) computes the maximum 2D distance between projections of the estimated and the ground-truth objects, and is more useful for application in virtual reality, where reprojection accuracy is more important
\end{itemize}
Additionally, for symmetric objects, \thirdchanges{we computed} MSSD and MSPD over all global symmetric transforms of the object and \thirdchanges{retaining the lowest error}. For each of the three metrics, average recall (AR) is calculated over different thresholds, above which an estimation is considered correct. 

We show extensive comparisons in Tables~\ref{tab:lmo},~\ref{tab:ycbv},~\ref{tab:tudl}. \secondchanges{On the overall benchmark, CosyPose~\cite{labbe2020cosypose} achieves state-of-the-art results and outperforms the other methods in most of the metrics. Our method, despite its broader scope, is competitive with the other methods in the 2020 BOP benchmark. In fact, for the YCB-V dataset, our method outperforms all other methods except CosyPose.} 

\secondchanges{Notably, our method has the second-highest VSD score across the three datasets. VSD calculates discrepancy on only the visible part of the objects. Because our optimization procedure weights the predicted keypoints by their confidence, and occluded keypoints often have lower confidence, our method favors poses that better match visible object-parts.}


\begin{table}[hbt!]
\centering
\small
\hspace{-3mm}
\tabcolsep=0.85mm
\begin{tabular}{@{}lccccc}
\toprule
\changes{Method} & test input & Avg.  & AR$_{mspd}$  & AR$_{mssd}$ & AR$_{vsd}$\\
\midrule
\changes{SSD-6D w/o ref~\cite{kehl2017ssd}}  & rgb &  0.139 & 0.285 & 0.083 & 0.047 \\
\changes{Pix2Pose~\cite{park2019pix2pose}}  & rgb &  0.363 & 0.550 & 0.307 & 0.233 \\
\changes{Leaping from 2D to 6D~\cite{liu2020leap}}    & rgb &  0.525 & 0.781 & 0.444 & 0.350 \\
\changes{CDPN~\cite{li2019cdpn}}     & rgb &  0.624 & \textbf{0.815} & \textbf{0.612} & 0.445 \\
\changes{CosyPose~\cite{labbe2020cosypose}}    & rgb & \textbf{0.633} & 0.812 & 0.606 & \textbf{0.480} \\
\changes{Ours} & rgb &  0.612 & 0.795 & 0.581 & 0.459 \\
\bottomrule
\end{tabular}
\caption{\changes{\textbf{Quantitative evaluation on LineMOD-Occluded~\cite{brachmann2014learning}.} Comparison to RGB-based 6-DoF pose estimation methods. Statistics are taken from the BOP benchmark~\cite{hodan2020bop}.}}
\label{tab:lmo}
\end{table}


\begin{table}[hbt!]
\centering
\small
\hspace{-3mm}
\tabcolsep=0.85mm
\begin{tabular}{@{}lccccc}
\toprule
\changes{Method} & test input & Avg.  & AR$_{mspd}$  & AR$_{mssd}$ & AR$_{vsd}$\\
\midrule
\changes{Pix2Pose~\cite{park2019pix2pose}}  & rgb &  0.457 & 0.571 & 0.429 & 0.372 \\
\changes{CDPN~\cite{li2019cdpn}}     & rgb &  0.532 & 0.631 & 0.570 & 0.396 \\
\changes{Leaping from 2D to 6D~\cite{liu2020leap}} & rgb &  0.543 & 0.687 & 0.499 & 0.443 \\
\changes{CosyPose~\cite{labbe2020cosypose}}    & rgb & \textbf{0.821} & \textbf{0.850} & \textbf{0.842} & \textbf{0.772} \\
\changes{Ours} & rgb & 0.618 & 0.700 & 0.605 & 0.549 \\
\bottomrule
\end{tabular}
\caption{\changes{\textbf{Quantitative evaluation on YCB-Video~\cite{calli2015ycb}.} Comparison to RGB-based 6-DoF pose estimation methods. Statistics are taken from the BOP benchmark~\cite{hodan2020bop}.}}
\label{tab:ycbv}
\end{table}

\begin{table}[hbt!]
\centering
\small
\hspace{-3mm}
\tabcolsep=0.85mm
\begin{tabular}{@{}lccccc}
\toprule
\changes{Method} & test input & Avg.  & AR$_{mspd}$  & AR$_{mssd}$ & AR$_{vsd}$\\
\midrule
\changes{Pix2Pose~\cite{park2019pix2pose}}  & rgb &  0.420 & 0.641 & 0.364 & 0.255 \\
\changes{\secondchanges{Leaping from 2D to 6D~\cite{liu2020leap}}} & rgb &  0.751 & 0.922 & 0.716 & 0.614 \\
\changes{CDPN~\cite{li2019cdpn}}     & rgb &  0.772 & 0.925 & 0.793 & 0.597 \\
\changes{CosyPose~\cite{labbe2020cosypose}}    & rgb & \secondchanges{\textbf{0.823}} & \secondchanges{\textbf{0.973}} & \secondchanges{\textbf{0.807}} & \secondchanges{\textbf{0.689}} \\
\changes{Ours} & rgb & 0.784 & 0.950 & 0.777 & 0.622 \\
\bottomrule
\end{tabular}
\caption{\changes{\textbf{Quantitative evaluation on TUD-Light~\cite{hodan2018bop}.} \secondchanges{Comparison to RGB-based 6-DoF pose estimation methods. Statistics are taken from the BOP benchmark~\cite{hodan2020bop}.}}}
\label{tab:tudl}
\end{table}

}

\subsection{Semi-Automated Data Labeling}
To apply keypoint-based pose estimation to new object classes, we must be able to rapidly collect images and annotate them with keypoints. We \changes{ demonstrate a significant reduction to the required} labeling effort with our projection based labeling tool.

To show the utility of our labeling tool, we answer several questions:
\begin{enumerate}
    \begin{minipage}{\textwidth}\item How well do the keypoints generated by the labeling tool match the keypoints labeled by human annotators?\end{minipage}
    \begin{minipage}{\textwidth}\item Do keypoint localization models trained with the annotations from the tool achieve good performance?\end{minipage}
    \begin{minipage}{\textwidth}\item Does the tool reduce the effort to label a set of images?\end{minipage}
    \secondchanges{\begin{minipage}{\textwidth}\item How does the keypoint-based refinement method compare with the model-based refinement method?\end{minipage}}
\end{enumerate}

\begin{figure}
    \begin{tabular}{P{4cm}P{4cm}P{4cm}P{4cm}}
    \multicolumn{4}{c}{                      \includegraphics[width=\linewidth]{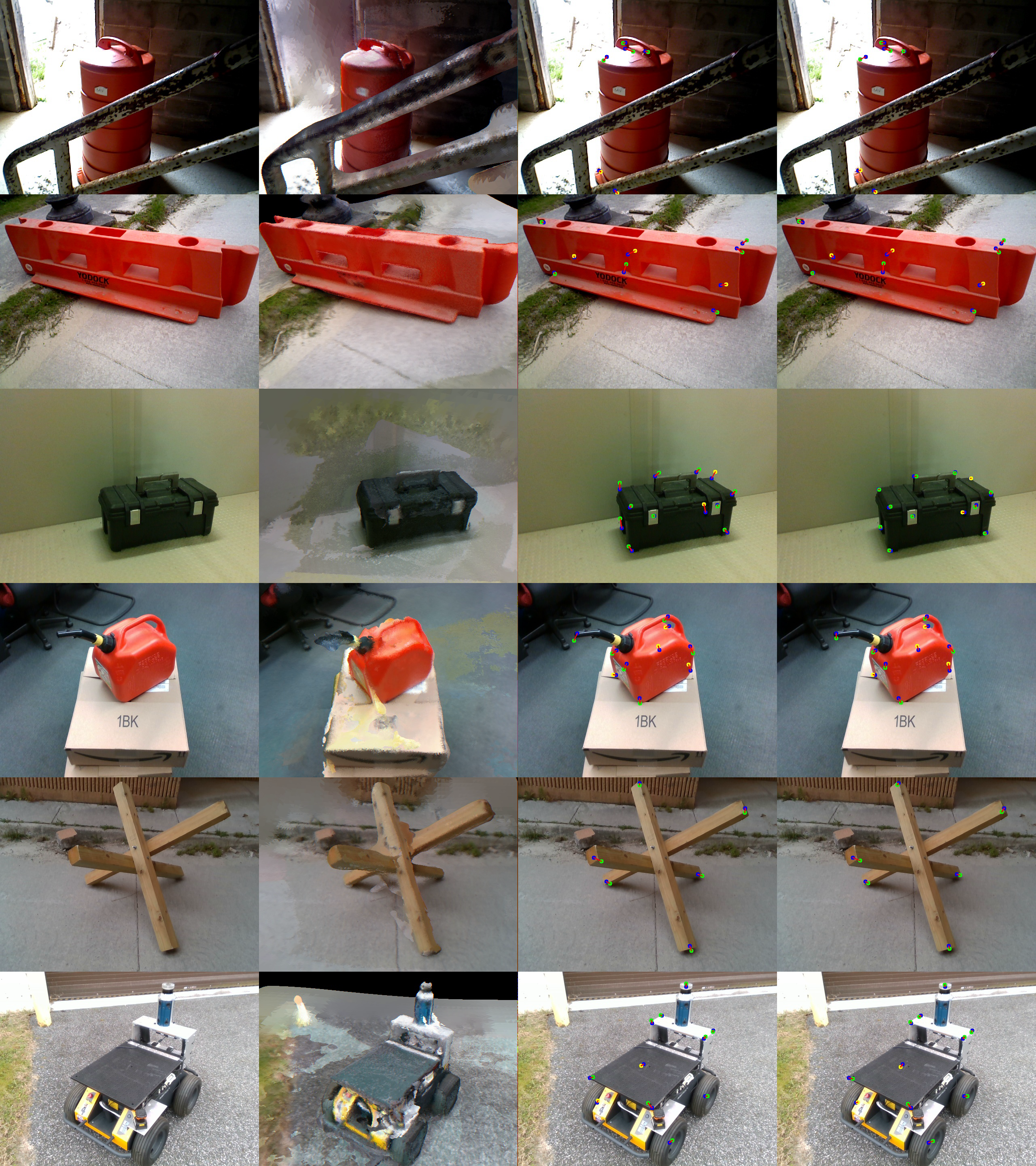}
    } \\
    \hspace{1cm} Input Image & \hspace{.25cm} Reconstructed Image & Projected Keypoints & \hspace{-.75cm} Refined Keypoints \\
    \end{tabular}
    \caption{Reconstructed images and \changes{semi-automated} keypoint projection. Keypoints are visualized as manual (blue dot), projected and visible (green dot), or projected and occluded (yellow dot). Red lines show the distances between manual keypoints and corresponding projected/refined keypoints.}
    \label{fig:keypoints_tiled}
\end{figure}

\begin{figure}
    \begin{tabular}{cccc}
        \includegraphics[width=45mm]{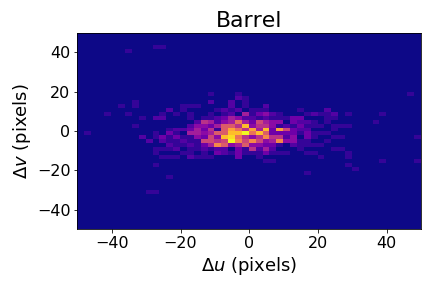} &      \includegraphics[width=45mm]{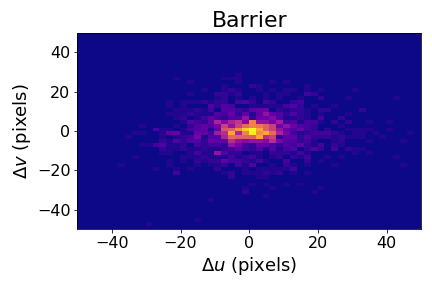} &     \includegraphics[width=45mm]{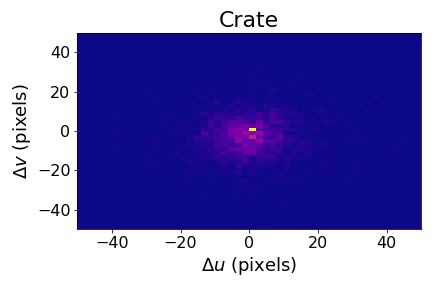} &
        \includegraphics[height=30mm,trim=0 -25mm 0 0]{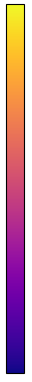} \\
        \includegraphics[width=45mm]{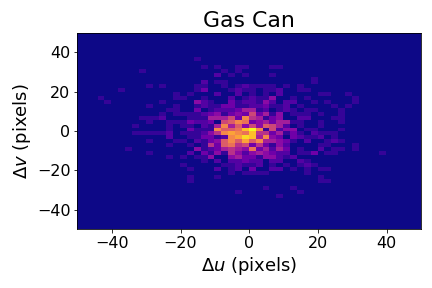} &      \includegraphics[width=45mm]{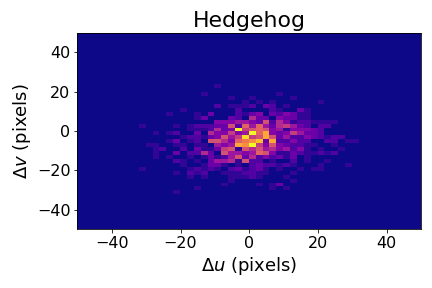} &     \includegraphics[width=45mm]{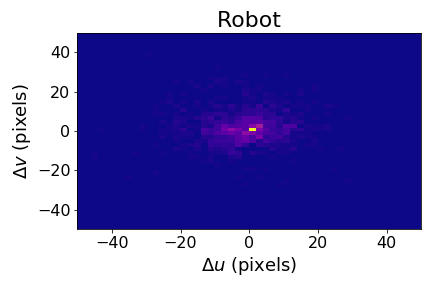} &
        \includegraphics[height=30mm,trim=0 -25mm 0 0]{figures/annotation/colorbar.png} \\
    \end{tabular}
    \caption{Pixel-wise distance between manual annotations and projected keypoints. The distance is a function of both the scene reconstruction pipeline accuracy as well as the manual annotation accuracy.}
    \label{fig:manual_automatic_distances}
\end{figure}

To answer these questions, a subset of the collected data was labeled by human annotators using a traditional annotation tool, whereby the user is prompted to select all object keypoints in randomly selected images. Histograms of the pixel-wise distances between manual and projected keypoint annotations for each class are given in Figure~\ref{fig:manual_automatic_distances}, showing approximately zero-mean distributions with different deviations for different objects. These differences are a function of both the accuracy of the reconstruction and camera pose estimation, as well as the accuracy of the manual annotations, which are not always reliable as ground truth surrogates, particularly in the case of occluded keypoints as illustrated in Figure~\ref{fig:keypoints_tiled}.



\begin{figure}
  \centering
  \begin{subfigure}[b]{0.45\textwidth}
    \includegraphics[width=\linewidth]{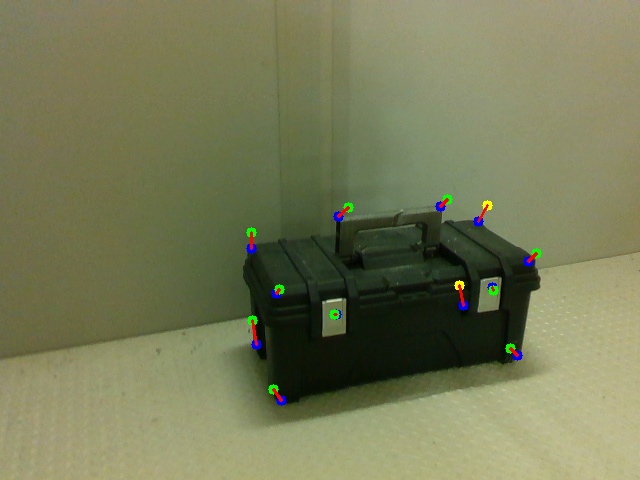}
    \caption{}
    \label{subfig:crate-projected}
  \end{subfigure}
  \begin{subfigure}[b]{0.45\textwidth}
    \includegraphics[width=\linewidth]{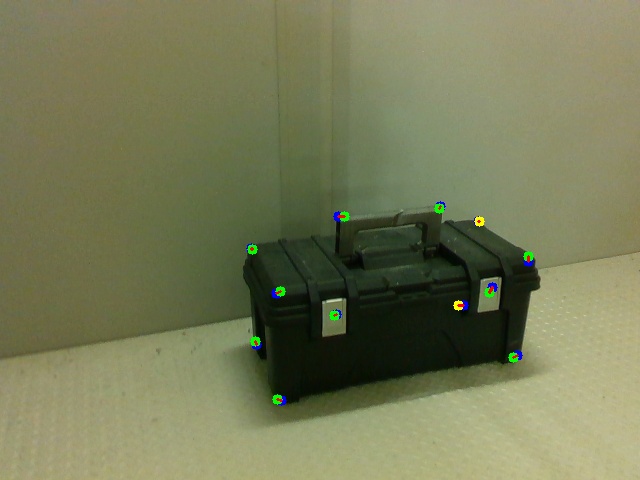}
    \caption{}
    \label{subfig:crate-refined}
  \end{subfigure}
  \caption{\changes{Demonstration of object-level refinement. Yellow circles indicate that a keypoint is occluded, green circles indicate visible keypoints, and blue circles indicate the manual annotations. For sequences with errors in sensor-trajectory estimation, in this case due to rapid camera motion with few available background features, naive projection from model to image frame, Figure~\ref{subfig:crate-projected}, exhibits poor alignment to manual annotations. After object-level refinement, Figure~\ref{subfig:crate-refined}, the keypoints are substantially improved. }}
  \label{fig:icp_improvement}
\end{figure}

\changes{
Table~\ref{tab:refinement_differences} compares the performance of the keypoint-level refinement and the object-level refinement against manually-annotated keypoints. For the visible case, where manual annotation is a better ground truth proxy, the object-level refinement outperforms the keypoint-level refinement. Therefore, we use the object-level refinement as the only refinement step for subsequent 
\thirdchanges{analysis, so following mentions of the process refer only to object-level refinement.}
Table~\ref{tab:occlusion_differences} compares the performance of naive projection to keypoint refinement, again using the projected pixel distance to manual annotations as an error metric. The results indicate that keypoint refinement may not be necessary for sequences with very good reconstructions, although we have identified some sequences that clearly benefit from refinement, such as shown in Figure~\ref{fig:icp_improvement}.
\secondchanges{Table~\ref{tab:refinement_differences} and Table~\ref{tab:occlusion_differences} also demonstrate the impact of visible versus occluded keypoints on manual annotation accuracy, with occluded keypoints always having higher distances between projected and manual annotations. While manual annotations for clearly visible keypoints, user error notwithstanding, can be considered as approximate ground truth, manual keypoints for occluded keypoints cannot. }
}

\begin{figure}
  \centering
  \begin{subfigure}[b]{0.45\textwidth}
    \includegraphics[width=\linewidth]{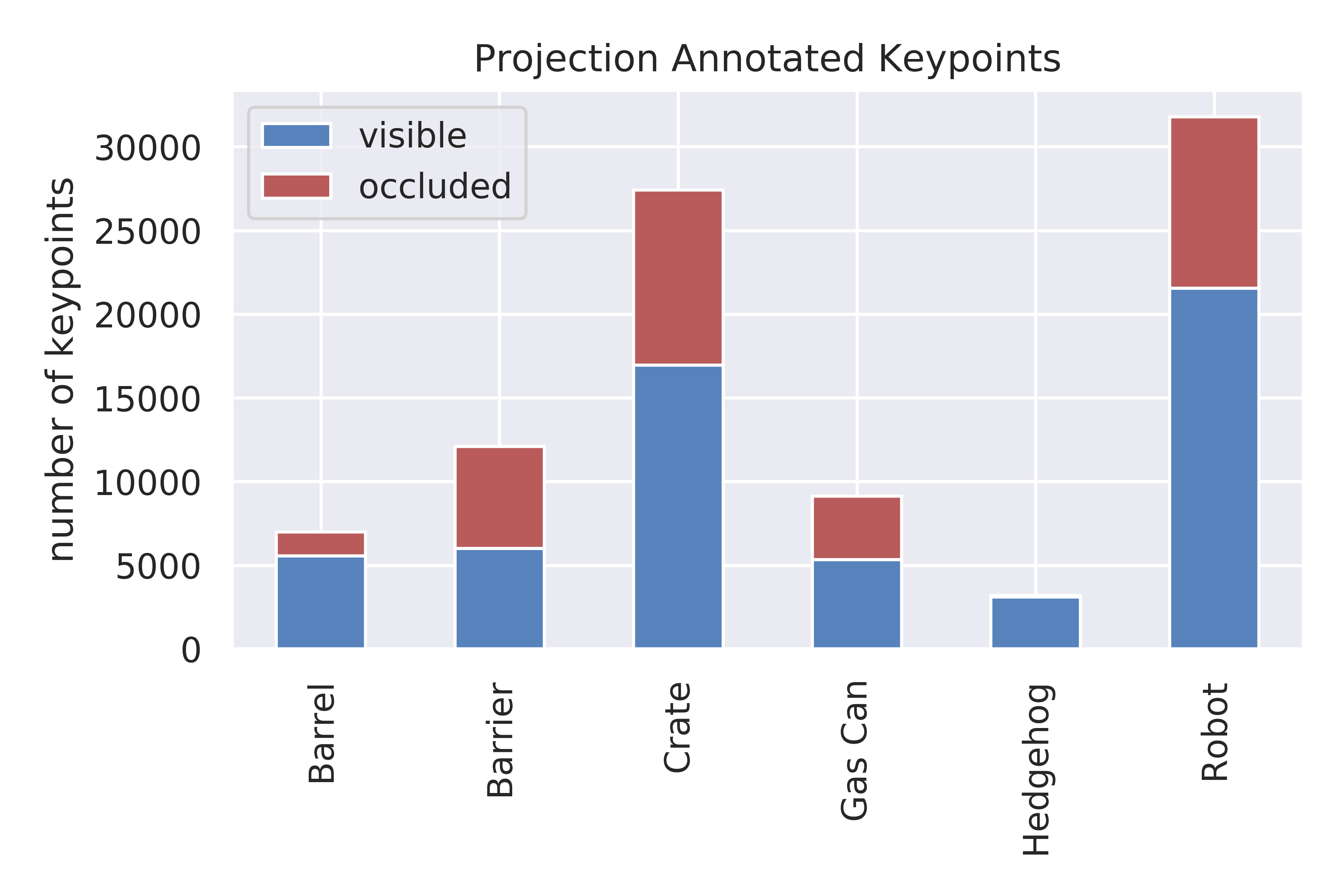}
    \caption{}
    \label{subfig:projected_vis_occ_stacked}
  \end{subfigure}
  \begin{subfigure}[b]{0.45\textwidth}
    \includegraphics[width=\linewidth]{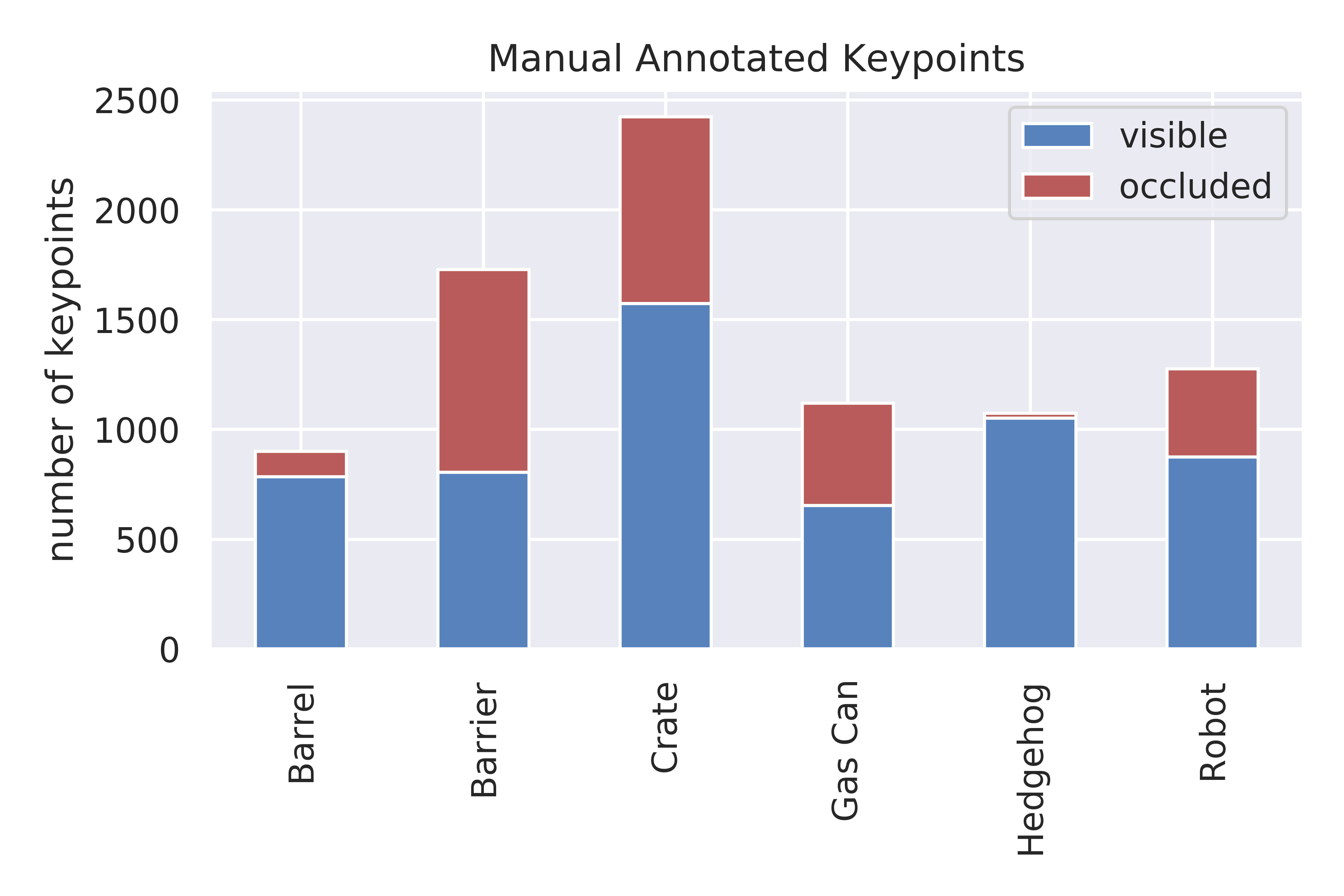}
    \caption{}
    \label{subfig:manual_vis_occ_stacked}
  \end{subfigure}
  \caption{\changes{Total number of visible versus occluded keypoints in the our dataset for all projected frames \ref{subfig:projected_vis_occ_stacked} and manually annotated frames \ref{subfig:manual_vis_occ_stacked} (note difference in scale between the plots). As a result of not performing full loops around symmetric objects such as the Barrel and Hedgehog, the number of occluded keypoints for symmetric objects is much smaller than for other objects.}}
  \label{fig:visible_occluded_stacked}
\end{figure}

For asymmetric objects, we ensured that each data collection contained a full loop around the object of interest, \thirdchanges{made up of a continuous sequence of images capturing the object from all angles}; however for symmetric objects (hedgehog and barrel), pose estimation is ambiguous about the axis of symmetry. Therefore, for such classes, each data sequence contains data for only one side of the object. As a consequence, these sequences have very few occluded keypoints, which results in similar distance results for both visible and occluded keypoints. \changes{Details on the distribution of occluded versus visible keypoints in the full dataset are shown in Figure~\ref{subfig:projected_vis_occ_stacked}, and the distribution of data used to compare manual and \changes{semi-automated} annotations are shown in Figure~\ref{subfig:manual_vis_occ_stacked}.
}



\begin{table}[t]
\begin{center}
\begin{tabular}{|c|c|c|c|c|c|c|}
  \hline
  & \multicolumn{2}{|c|}{All} & \multicolumn{2}{|c|}{Visible} & \multicolumn{2}{|c|}{Occluded}\\
  \hline
   & Keypoint & Object & Keypoint & Object  & Keypoint & Object \\
  \hline\hline
  Barrel & $11.26\pm0.3$ & $\mathbf{11.05\pm0.3}$ & $11.04\pm0.3$ & $\mathbf{10.77\pm0.3}$ & $\mathbf{12.78\pm1.0}$ & $12.96\pm1.0$ \\ \hline
  Barrier & $\mathbf{11.90\pm0.2}$ & $11.92\pm0.2$ & $\mathbf{9.33\pm0.2}$ & $9.47\pm0.2$ & $14.14\pm0.4$ & $\mathbf{14.07\pm0.4}$ \\ \hline
  Crate & $11.16\pm0.2$ & $\mathbf{11.14\pm0.2}$  & $9.27\pm0.2$ & $\mathbf{9.07\pm0.1}$ & $\mathbf{14.66\pm0.4}$ & $14.98\pm0.4$ \\ \hline
  Gas Can & $12.58\pm0.2$ & $\mathbf{12.19\pm0.2}$ & $11.06\pm0.3$ & $\mathbf{10.42\pm0.3}$ & $14.73\pm0.4$ & $\mathbf{14.67\pm0.4}$  \\ \hline
  Hedgehog & $\mathbf{11.64\pm0.2}$ & $11.75\pm0.2$ & $\mathbf{11.64\pm0.2}$ & $11.71\pm0.2$ & $\mathbf{11.79\pm1.1}$ & $13.20\pm1.2$ \\ \hline
  Robot & $10.43\pm0.2$ & $\mathbf{10.02\pm0.2}$ & $8.48\pm0.2$ & $\mathbf{7.96\pm0.2}$ & $14.67\pm0.6$ & $\mathbf{14.52\pm0.6}$ \\ \hline
\end{tabular}
\end{center}
\caption{\changes{Evaluation of 
\thirdchanges{our} refinement techniques, keypoint-level (feature matching with jump edge refinement) and object-level refinement (ICP). Error is measured as the $\ell_2$ distances in pixels for the refinement methods with respect to manual annotations.  The object-level context used by ICP to refine all keypoints performs better than the per-keypoint adjustment of feature matching with jump edge refinement, so we adopt the object-level method as our default refinement technique.}
} \label{tab:refinement_differences}
\end{table}

\begin{table}[t]
\begin{center}
\begin{tabular}{|c|c|c|c|c|c|c|}
  \hline
  & \multicolumn{2}{|c|}{All} & \multicolumn{2}{|c|}{Visible} & \multicolumn{2}{|c|}{Occluded}\\
  \hline
  & Projected & Refined & Projected & Refined & Projected & Refined \\
  \hline\hline
  Barrel & $11.29\pm0.3$ & $\mathbf{11.05\pm0.3}$ & $11.07\pm0.3$ & $\mathbf{10.77\pm0.3}$ & $\mathbf{12.78\pm1.0}$ & $12.96\pm1.0$ \\ \hline
  Barrier & $12.07\pm0.2$ & $\mathbf{11.92\pm0.2}$ & $9.71\pm0.2$ & $\mathbf{9.47\pm0.2}$ & $14.14\pm0.4$ & $\mathbf{14.07\pm0.4}$ \\ \hline
  Crate & $\mathbf{10.87\pm0.2}$ & $11.14\pm0.2$  & $\mathbf{8.82\pm0.1}$ & $9.07\pm0.1$ & $\mathbf{14.66\pm0.4}$ & $14.98\pm0.4$ \\ \hline
  Gas Can & $12.44\pm0.2$ & $\mathbf{12.19\pm0.2}$ & $10.82\pm0.3$ & $\mathbf{10.42\pm0.3}$ & $14.73\pm0.4$ & $\mathbf{14.67\pm0.4}$  \\ \hline
  Hedgehog & $\mathbf{11.60\pm0.2}$ & $11.75\pm0.2$ & $\mathbf{11.59\pm0.2}$ & $11.71\pm0.2$ & $\mathbf{11.79\pm1.1}$ & $13.20\pm1.2$ \\ \hline
  Robot & $\mathbf{9.90\pm0.2}$ & $10.02\pm0.2$ & $\mathbf{7.72\pm0.2}$ & $7.96\pm0.2$ & $14.67\pm0.6$ & $\mathbf{14.52\pm0.6}$ \\ \hline
\end{tabular}
\end{center}
\caption{Visible versus occluded disparities for manual annotations.  $\ell_2$ distances in pixels for both the \changes{semi-automated} keypoint projection method as well as \changes{object-level} refinement are given with respect to manual annotations.  We show the difference for all the keypoints, for only the keypoints that are visible to the camera, and for only the keypoints that are occluded because they are on the far side of the object. There is a clear difference between the offsets for visible versus occluded keypoints, demonstrating the difficulty of accurately annotating occluded keypoints by hand.
} \label{tab:occlusion_differences}
\end{table}


Next, we analyzed the performance of the keypoint localization model trained with keypoints from different sources.
We trained three keypoint localization models for each class: (i) with the manual annotations, (ii) with the \changes{semi-automated} annotations, and (iii) with the refined annotations.
We then evaluate the performance of these models on a held-out set of images with manually labeled keypoints, as shown in Table~\ref{tab:distance_from_test_keypoints}.
Keypoint localization models trained with our refined keypoints consistently achieved performance comparable to the model trained with the manually labeled keypoints, and had a lower error on four out of six object classes.
The model trained with our refined keypoints perform worse only on the barrel and the hedgehog,  the only two objects with rotational symmetry.

\begin{figure}
  \centering
  \begin{subfigure}[b]{0.49\textwidth}
    \includegraphics[width=\linewidth]{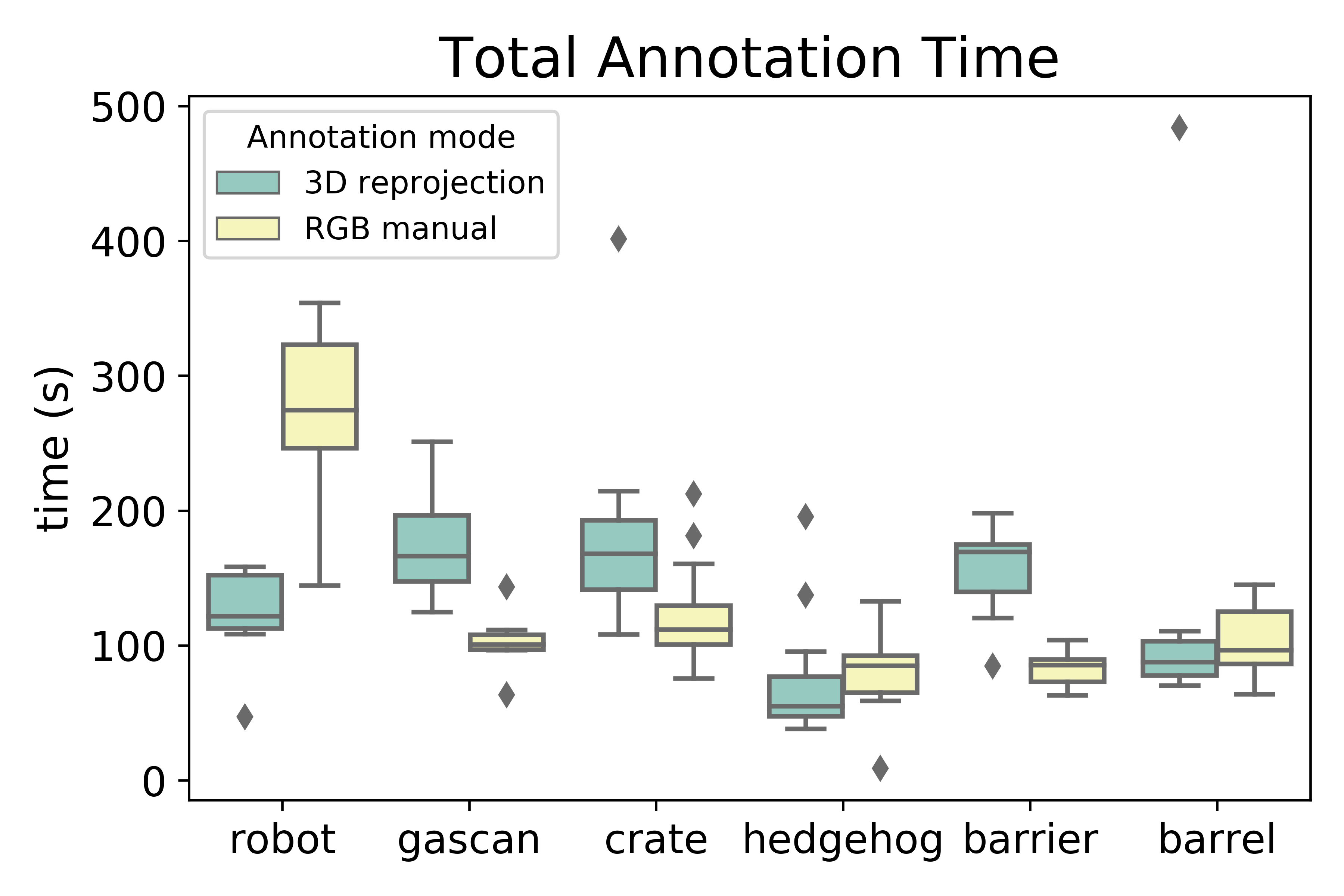}
    \caption{}
    \label{subfig:annotation_times}
  \end{subfigure}
  \begin{subfigure}[b]{0.49\textwidth}
    \includegraphics[width=\linewidth]{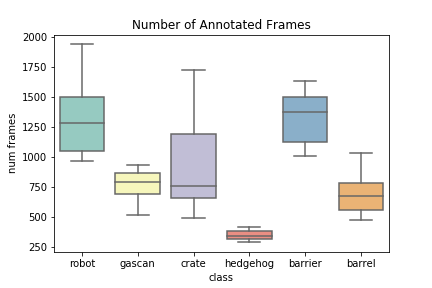}
    \caption{}
    \label{subfig:num_frames}
  \end{subfigure}
  \caption{Analysis between manual versus \changes{semi-automated} projection labeling. For each collected sequence for a given class, human annotation times (a) and sequence length (b) are shown. A human annotator is able to annotate a 3D model in approximately the same time it takes to annotate 100 individual image frames. The average number of annotated frames for each class is between one and two orders of magnitude greater for the same effort when using our annotation tool.}
  \label{fig:runtimes and number of frames}
\end{figure}

We conclude with an analysis of annotation effort required.
Measurements of the required time are shown in Figure~\ref{fig:runtimes and number of frames}. 
\thirdchanges{As expected, using semi-automated projection allows the user to label an order of magnitude more frames, within the same quantity of time, as they would be able to label with a standard annotation tool.}
The annotations for each object class, with approximately 100 total frames per class labeled using standard image annotation, take about the same time as it takes to annotate a 3D reconstructed model, \changes{which we project to label thousands of images per class}.
For this analysis, we subsampled the number of projected frames by a factor of 10; therefore the true number of potentially annotated frames using our method is even higher than shown. We chose to subsample the output frames to reduce the total projection time as well as to avoid training with essentially redundant camera angles.


\begin{table}[t]
\begin{center}
\begin{tabular}{|c|c|c|c|}
  \hline
   & Manual & Projected & Refined  \\
  \hline\hline
  Barrel & $\mathbf{6.7\pm0.7}$ & $7.0\pm0.8$ & $8.9\pm0.9$\\ \hline
  Barrier & $14.7\pm1.0$ & $15.1\pm1.1$ & $\mathbf{14.1\pm1.1}$\\ \hline
  Crate & $6.5\pm0.6$ & $22.9\pm0.96$ & $\mathbf{5.6\pm0.6}$\\ \hline
  Gascan & $13.4\pm0.8$ & $\mathbf{5.7\pm0.56}$ & $8.4\pm0.8$ \\ \hline
  Hedgehog & $\mathbf{1.6\pm0.1}$ & $2.0\pm0.1$ & $1.9\pm0.1$ \\ \hline
  Robot & $7.2\pm0.7$ & $5.0\pm0.6$ & $\mathbf{3.0\pm0.3}$  \\ \hline
\end{tabular}
\end{center}
\caption{Keypoint prediction performance on test data. $\ell_2$ Distance of predicted keypoints locations in pixel space from manually labeled test set.  The models trained with data from the refined labeling pipeline outperform the models trained with the manually labeled data at predicting the locations of manually labeled keypoints on four out of six object classes, while requiring less labeling effort.}
\label{tab:distance_from_test_keypoints}
\end{table}

\section{Conclusion}

In this paper, we 
\thirdchanges{presented} an efficient method to estimate the continuous 6-DoF pose of an object from a single RGB image. Capitalizing on the robust, semantic keypoint predictions provided by a state-of-the-art convnet, we proposed a pose optimization scheme that fits a deformable shape model to the 2D keypoints and recovers the object's 6-DoF pose. To ameliorate the effect of false detections, 
\thirdchanges{our pose optimization scheme integrates heatmap values, which reflect predictive confidence, to model each detection's certainty.}
Both the weak perspective and the full perspective cases were investigated. 
Moreover, we incorporated a technique for generating data from unlabeled videos, enabling \changes{semi-}automatic labeling, such that our pipeline can be trained with minimal effort during the annotation stage.
%
\changes{The experimental validation includes evaluation on instance-based pose estimation on the LineMOD-Occluded, YCB-Video, and TUD-Light datasets as well as evaluation on category-level pose estimation on the PASCAL3D+ dataset.}
Additionally, our method is accompanied by an efficient implementation with a running time under 0.3 seconds, making it a good fit for near-realtime robotics applications.


\subsubsection*{Acknowledgments}

This research  was  sponsored  by  the  U.S. Army  Research  Laboratory  (ARL)  under Cooperative Agreement W911NF-10-2-0016. The views and conclusions contained in this document are those of the authors and should not be interpreted as representing the official policies, either expressed or implied, of ARL or the U.S. Government. We would like to thank Shiyani Patel for her help with data collection and analysis, as well as RCTA program leaders, in particular Stuart Young, Dilip Patel, Dave Baran, and Geoff Slipher.

\bibliographystyle{apalike}
\bibliography{jfrExampleRefs}

\end{document}